\documentclass[runningheads]{llncs}

\usepackage{eccv}

\usepackage{eccvabbrv}

\usepackage{graphicx}
\usepackage{booktabs}
\usepackage{dsfont}

\usepackage[accsupp]{axessibility}  %

\usepackage{multibib}
\newcites{A}{Appendix References}

\usepackage{etoolbox}
\usepackage{comment}

\newbool{dummyappendix}
\boolfalse{dummyappendix} %

\ifbool{dummyappendix}{
    \excludecomment{appendixblock}
}{
    \includecomment{appendixblock}
}

\usepackage[pagebackref]{hyperref}

\usepackage{orcidlink}
\usepackage{enumitem}

\usepackage[capitalize,noabbrev]{cleveref}
\usepackage{algpseudocode}
\usepackage{algorithm}
\usepackage{xparse}
\makeatletter
\NewDocumentCommand{\LeftComment}{s m}{%
  \Statex \IfBooleanF{#1}{\hspace*{\ALG@thistlm}}\(\triangleright\) #2}
\makeatother\makeatletter

\usepackage{forloop}
\newcounter{ct}
\newcommand{\markdent}[1]{\forloop{ct}{0}{\value{ct} < #1}{\hspace{\algorithmicindent}}}
\newcommand{\markcomment}[1]{\Statex\markdent{#1}}

\usepackage{microtype}
\usepackage{graphicx}
\usepackage{subcaption}
\usepackage{booktabs} %

\usepackage{hyperref}
\usepackage{fontawesome5}
\usepackage{amsmath}
\usepackage{amssymb}
\usepackage{mathtools}

\usepackage{nicefrac}
\usepackage[misc]{ifsym}
\usepackage{multirow}

\usepackage{tikz}
\usetikzlibrary{positioning, arrows.meta, fit, calc, backgrounds}
\usepackage{adjustbox}

\newcommand{\bfparagraph}[1]{\paragraph{#1}}%

\usepackage{tikz}
\usetikzlibrary{shapes, arrows.meta, positioning, calc, backgrounds, fit}

\usepackage[capitalize,noabbrev]{cleveref}

\DeclareMathOperator*{\argmin}{arg\,min}
\crefname{definition}{Definition}{Definitions}
\crefname{equation}{Equation}{Equations}
\crefname{table}{Table}{Tables}
\crefname{figure}{Figure}{Figures}
\crefname{appendix}{Appendix}{Appendices}

\begin{document}

\title{On the Faithfulness of Post-Hoc Concept Bottleneck Models} 

\titlerunning{On the Faithfulness of Post-Hoc Concept Bottleneck Models}

\author{Laines Schmalwasser\inst{1,2}\orcidlink{0009-0006-1120-1299} \and
Jan Blunk\inst{2,3}\thanks{Work done while the author was at Computer Vision Group Jena.}%
\orcidlink{0009-0001-7037-3545} \and
Niklas Penzel\inst{2}\orcidlink{0000-0001-8002-4130} \and 
Julia Niebling\inst{1}\orcidlink{0000-0001-5413-2234} \and
Joachim Denzler\inst{2}\orcidlink{0000-0002-3193-3300}
}

\authorrunning{L.~Schmalwasser et al.}

\institute{Institute of Data Science, German Aerospace Center, Jena, Germany \and
Computer Vision Group Jena, Friedrich Schiller University Jena, Germany \and
GEOMAR Helmholtz Centre for Ocean Research Kiel, Germany\\
\textsuperscript{\Letter} Correspondence to \email{laines.schmalwasser@dlr.de}}

\maketitle

\begin{abstract}
Human decision-making interprets the world through high-level concepts, such as recognizing a bird by its belly color.
To bridge the gap between opaque deep learning representations and human understanding, Post-Hoc Concept Bottleneck Models (post-hoc CBMs) project latent features onto interpretable concept spaces using auxiliary datasets or vision-language models. 
However, relying on target task accuracy as the primary measure of post-hoc CBM success obscures whether the learned concepts are semantically meaningful or merely predictive artifacts.
For example, random concept projections can achieve competitive accuracy despite being semantically meaningless.
In this work, we analyze the learned projections directly and identify two failure cases:
First, for concept projections learned from auxiliary data, covariate shifts can lead to unfaithful concept representations for the target task.
In particular, we provide an upper bound on the error introduced by this shift.
Second, systematic label noise in surrogate concept labels generated by vision-language models leads to unfaithful projections. 
After formalizing these failure modes, we introduce novel metrics that decouple concept faithfulness from predictive accuracy.
Our empirical results across real-world and synthetic benchmarks confirm that these metrics identify unfaithful behaviors that standard accuracy-based evaluation fails to detect\footnote{Project page: \url{https://posthoc-cbm-faithfulness.github.io/}}. %
\end{abstract}

\section{Introduction} 
Human understanding is fundamentally built on concepts~\cite{barsalou1999perceptual,gardenfors2004conceptual,murphy2004big}.
Rather than analyzing raw sensory data, we identify objects through high-level attributes, such as recognizing a specific bird species by its ``yellow belly'' or ``black crown.''
In contrast, deep neural networks learn opaque, high-dimensional representations that often lack semantic clarity~\cite{lipton2018interpretability,koh2020concept,bau2017network}.
To bridge this gap and enforce interpretability, Concept Bottleneck Models (CBMs) were introduced~\cite{koh2020concept,alukaev2023cross, chauhan2023interactive,kazmierczak2024clip,moayeri2023text,tan2024explain,vandenhirtz2024stochastic,almudevar2025there,galliamov2025concepts}.
CBMs constrain the model to first predict a set of human-understandable concepts from the input, and then base the final prediction solely on them, thereby promising transparent decision-making~\cite{jacovi2020towards,mahinpei2021promises,margeloiu2021concept}.

However, standard CBMs require dense concept annotations for the specific downstream task, which are rarely available in practice~\cite{oikarinen2023label,koh2020concept,zarlenga2022concept}.
Post-Hoc Concept Bottleneck Models (post-hoc CBMs) circumvent this by freezing a pretrained feature extractor and learning only a lightweight projection from its internal representations to a concept space and final classifier~\cite{yuksekgonul2023posthoc,oikarinen2023label,srivastava2024vlg} (we use post-hoc CBM to denote this general model class and PCBM for the concrete implementation in \cite{yuksekgonul2023posthoc}).
Crucially, post-hoc CBMs tackle the need for in-domain concept labels via two strategies:  (1) \emph{Auxiliary Concept Datasets}, where the projection is learned on a separate, richly annotated dataset (e.g., Broden~\cite{bau2017network}) before application to the target task \cite{yuksekgonul2023posthoc}; or (2) \emph{Surrogate Labels from VLMs}, where a Vision-Language Model (e.g., CLIP~\cite{radford2021learning}) generates surrogate concept labels~\cite{oikarinen2023clip,oikarinen2023label,srivastava2024vlg,sampat2024help}.
Both drastically reduce the annotation burden while maintaining competitive downstream classification accuracy~\cite{yuksekgonul2023posthoc,oikarinen2023label}.

\begin{figure}[t]
    \centering
    \resizebox{\textwidth}{!}{\input{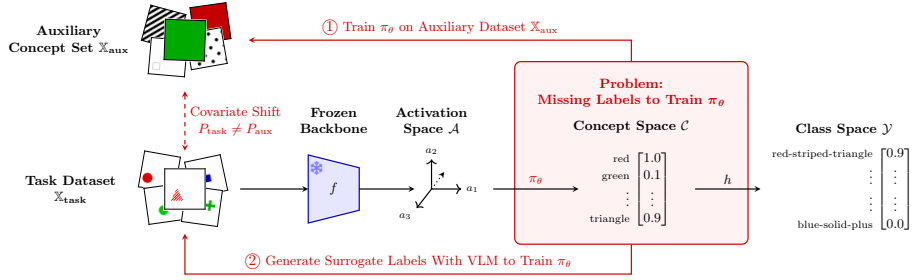}}
    \caption{
    In the post-hoc CBM setup \cite{yuksekgonul2023posthoc,oikarinen2023label,srivastava2024vlg}, inputs are mapped to activations using some frozen backbone model $f$. 
    Then, a learned concept projection $\pi_\theta$ extracts concepts before a final classifier $h$ predicts a class based on these concepts.
    Unfortunately, concept labels in the downstream task domain are often unavailable.
    Thus, \textcircled{1} auxiliary concept sets (e.g., Broden \cite{bau2017network}) or \textcircled{2} surrogate VLM labels are used \cite{yuksekgonul2023posthoc,oikarinen2023label,srivastava2024vlg}. 
    We study these approaches and identify two reasons for potential unfaithfulness of $\pi_\theta$: 
    \textcircled{1} a covariate shift of the concept set, and \textcircled{2} systematic surrogate label errors.
    }
    \label{fig:teaser}
\end{figure}

Unfortunately, evaluating downstream accuracy provides limited insight into the quality of the learned concept projection.
Prior work empirically shows that random concepts achieve high performance~\cite{midavaine2024reproducibility}.
In \cite{makonnen2025measuring}, this phenomenon is connected to the Johnson-Lindenstrauss (JL) lemma~\cite{johnson1984extensions}, with the expected error approaching zero as random projections increase \cite{srivastava2024vlg}.
We expand these insights via the smooth manifold JL lemma \cite{baraniuk2009random}, based on previous findings that neural activations tend to lie on low-dimensional manifolds \cite{ansuini2019intrinsic,pope2021intrinsic}.
We show that under certain assumptions, the original activations can be reconstructed from a modest number of random projections, resulting in high downstream performance.

Consequently, we must investigate the concept projection itself to evaluate whether it faithfully extracts the intended concepts on the target distribution.
Here, we analyze the two primary post-hoc CBM training mechanisms (\Cref{fig:teaser}).
First, for auxiliary concept datasets, we identify \emph{covariate shift} as the primary source of unfaithfulness.
We demonstrate that even if a concept's semantic definition is identical across domains, a geometric shift in the feature space can invalidate the learned projection on the downstream task. 
We formalize this by providing an upper bound for the error introduced by this shift based on \cite{BenDavid.2010} and propose an approximation of it as a practical measure of faithfulness.
Second, for methods using VLM-based surrogate labels, the primary problem is \emph{label noise}.
However, our analysis shows that unfaithfulness is not solely caused by noise magnitude, but by systematic errors.
In other words, random mistakes can cancel each other out, but the surrogate labeling function leads to unfaithfulness when mistakes are consistent, e.g., always predicting ``sky'' and ``clouds'' together.
To detect this, we measure the surrogate-label error and propose a metric that explicitly quantifies the corresponding systematicity.

We empirically validate our theoretical findings on standard benchmarks (CUB-200~\cite{welinder2010cub} and CIFAR-10/100~\cite{krizhevsky2009cifar}) and the synthetic Elements~\cite{nicolson2025explaining} dataset, which provides known ground-truth concepts.
Our evaluation includes a range of post-hoc CBMs across standard backbones representing the two main training paradigms: those using auxiliary datasets \cite{yuksekgonul2023posthoc} (e.g., Broden~\cite{bau2017network}) and those using surrogate labels generated by VLMs (LFCBM~\cite{oikarinen2023label} and VLG-CBM~\cite{srivastava2024vlg} based on CLIP~\cite{radford2021learning}, DINOv3~\cite{simeoni2025dinov3} and Grounding DINO \cite{liu2024grounding}).
We first show that random concept projections can achieve competitive accuracy, confirming that downstream accuracy is an insufficient metric for meaningful representations.
Then, we directly evaluate the learned concept projections using available ground truth \cite{nicolson2025explaining}, apply our proposed faithfulness metrics, and demonstrate that they successfully identify failures caused by covariate shifts and systematic errors, offering a practical tool for evaluating post-hoc CBMs beyond predictive performance.

Our contributions can be summarized as follows:
\begin{enumerate}
    \item We show that under certain assumptions, random concept projections can achieve competitive performance.
    \item We identify and formalize two critical sources of unfaithfulness in the concept projections of post-hoc CBMs: \emph{covariate shifts} when learning with auxiliary concept sets and \emph{systematic label noise} in surrogate supervision.
    \item We propose a set of novel faithfulness metrics that evaluate concept alignment independently of the predictive performance, revealing specific failure cases. 
\end{enumerate}

\section{Related Work}
\label{sec:related-work}

post-hoc CBM evaluations are complicated by information leakage \cite{mahinpei2021promises,margeloiu2021concept,havasi2022addressing,sun2024eliminating,schoen2025measuring}, where models bypass semantic concepts to base predictions on raw backbone activations.
Notably, \cite{schoen2025measuring} links this phenomenon to the Johnson-Lindenstrauss lemma \cite{johnson1984extensions}, showing that sufficient random projections preserve pairwise distances for finite sets.
We note that the manifold Johnson-Lindenstrauss lemma \cite{baraniuk2009random} implies this holds for the complete activation manifold under the manifold hypothesis \cite{ansuini2019intrinsic,pope2021intrinsic}. 
Related to this, \cite{srivastava2024vlg} studies the expected error of random projections for final predictions. 
Based on these findings, we emphasize the need to evaluate learned concept projections directly.
Crucially, the faithfulness issues we identify persist even in the absence of information leakage.

Other works similarly study CBM and post-hoc CBM limitations \cite{zarlenga2022concept,ramaswamy2023overlooked,huang2024concept,luyten2024theoretical} and concept faithfulness \cite{lai2024faithful,kumar2025measuring}.
While \cite{kumar2025measuring} analyzes the faithfulness of unsupervised concept explanations, it explicitly excludes post-hoc CBMs (our focus).
For surrogate labels, \cite{lai2024faithful} highlights flaws in GPT-3-derived concept sets \cite{brown2020language} for label-free CBMs \cite{oikarinen2023label}.
We generalize this by providing an alternative mechanism to analyze surrogate-label errors in VLM-based post-hoc CBMs.
In concept embedding models \cite{zarlenga2022concept}, the concept bottleneck is constructed from two learned embeddings per concept to address the accuracy-interpretability trade-off.
In contrast to our evaluation metrics, \cite{huang2024concept} studies CBM faithfulness by matching input regions to concepts.
Finally, \cite{ramaswamy2023overlooked} empirically observes issues caused by distribution shifts in concept datasets.
We provide an upper bound on the risk of post-hoc CBMs trained on auxiliary data, approximating the resulting generalization error.
In contrast, \cite{luyten2024theoretical} bounds the CBM risk on downstream tasks using the underlying backbone.

\section{Method}
\label{sec:method}

We consider a standard deep learning setting where a fixed backbone $f: \mathcal{X} \to \mathcal{A} \subseteq \mathbb{R}^d$ maps inputs to a high-dimensional latent activation space.
While a classifier $g: \mathcal{A} \to \mathcal{Y}$ trained on top of $f$ may achieve high performance, its decision-making process within the high-dimensional space $\mathcal{A}$ remains opaque.
Post-Hoc Concept Bottleneck Models (post-hoc CBMs)~\cite{yuksekgonul2023posthoc,oikarinen2023label,srivastava2024vlg} aim to make this process interpretable by introducing an intermediate concept bottleneck while holding $f$ fixed.
Specifically, they learn a concept projection $\pi_{\theta}: \mathcal{A} \to \mathcal{C}$, parameterized by $\theta$, that maps activations to a $K$-dimensional space $\mathcal{C}~\subseteq~\mathbb{R}^K$ of human-understandable concepts (typically $K \ll d$).
A subsequent classifier $h: \mathcal{C} \to \mathcal{Y}$ (e.g., a sparse linear layer) predicts target classes based on these concept scores, yielding the final model $h \circ \pi_\theta \circ f$.

To construct the bottleneck layer, we initially assume access to a training set $\mathbb{X}_{\mathrm{task}} = \{x_i, c_i\}_{i=1}^N$ sampled i.i.d. from a task distribution $P_{\mathrm{task}}(\mathcal{X}, \mathcal{C})$ over inputs and concept vectors, allowing us to optimize $\theta$ to recover these task-relevant concepts.
\begin{definition}\label{def:faithful}
We define a concept projection $\pi_{\theta}$ as \emph{faithful} if it minimizes the expected risk over the true task distribution $P_\mathrm{task}(x,c)$:
\begin{equation}\label{eq:faithful}
\argmin_\theta \mathbb{E}_{(x,c) \sim P_\mathrm{task}} \left[ \mathcal{L}(\pi_\theta(f(x)), c)\right],
\end{equation}
for some loss function $\mathcal{L}$ measuring the alignment of the predicted concepts $\pi_\theta(f(x))$ and the ground truth concepts $c$ for an input $x$.
\end{definition}

In this work, we study the post-hoc concept bottleneck framework from this perspective of concept faithfulness, i.e., whether $\pi_\theta$ extracts meaningful concepts after \Cref{def:faithful}.
A faithful projection ensures $\pi_\theta(f(x)) = \mathbb{E}_{P_\mathrm{task}}[c\mid x]$, recovering the intended concepts in expectation.
Note that our formalization of faithfulness differs from similar notions in information leakage, e.g., \cite{margeloiu2021concept,makonnen2025measuring,schoen2025measuring}, and aligns more closely with the informal version in \cite{zarlenga2022concept}.

First, we confirm that evaluating the predictive performance of $h$ is insufficient to assess the faithfulness of the post-hoc CBM in \Cref{sec:method-downstream}.
After establishing that downstream accuracy does not imply a faithful concept projection, we formalize the learning problem of $\pi_{\theta}$ in \Cref{sec:method-concept-projection} to derive direct faithfulness measures. 
We identify two primary sources of unfaithfulness: (1) \emph{covariate shifts} from auxiliary training datasets, which we quantify via an upper bound on the generalization error; and (2) \emph{systematic label noise} from VLM-generated surrogate concept labels.
For surrogate labels, we show that unfaithfulness requires non-random error structures rather than magnitude alone, and derive a metric to explicitly quantify this systematicity.

\subsection{Downstream Performance and Faithfulness}
\label{sec:method-downstream}
To demonstrate that high post-hoc CBM performance does not imply semantic alignment, we study random projections $\pi_\theta$ without semantic meaning, 
e.g., independently sampled standard-normal coefficients in a linear layer.
Previous work empirically shows that PCBMs \cite{midavaine2024reproducibility} using such $\pi_\theta$ can achieve high downstream performance.
In \cite{makonnen2025measuring}, the authors note that such projections are geometry-preserving with high probability for finite sets of samples under the Johnson-Lindenstrauss (JL) lemma \cite{johnson1984extensions} if the embedding dimension $K$ is large enough.
In other words, the complete backbone information is encoded into the concept activation, which is related to the area of information leakage \cite{margeloiu2021concept,havasi2022addressing,makonnen2025measuring,schoen2025measuring}.

Here, we first link this to the manifold hypothesis, then derive a concrete mechanism to achieve high downstream performance under additional assumptions.
In particular, consider the manifold hypothesis of deep learning \cite{ansuini2019intrinsic,pope2021intrinsic}, i.e., that the activations in neural networks lie on or close to a manifold with an intrinsic dimension of $m \ll d$.
If this hypothesis holds, then the manifold version of the JL lemma (Theorem 3.1 in \cite{baraniuk2009random}) governs the behavior under random projections.
Crucially, it establishes a connection between certain characteristics of this activation manifold and the number of random projections needed to preserve pairwise distances.
We make this explicit in \Cref{apx:jl-lemma}.
Nevertheless, if geometry is preserved, the original activations can be recovered \cite{baraniuk2009random}.
Thus, for a sufficiently complex classifier $h$ (e.g., an MLP \cite{hornik1989multilayer}) it is possible to learn $g \circ \pi_\theta^{-1}$.
In such a case, the meaningless concepts by the random $\pi_\theta$ would be used to recover the original activations, leading to high downstream performance.
To make this argument more explicit, we next derive a concrete solution for $h$ under an additional linearity assumption.

\bfparagraph{Constructive Proof for a Linear Activation Subspace.}
While the discussion under the manifold hypothesis posits a possible solution for $h$, we now derive a constructive solution under the assumption that the activation manifold is a linear subspace of $\mathbb{R}^d$.
Our concrete assumptions are:

\begin{enumerate}[label=(\arabic*)]
    \item \textbf{Linear Subspace:} The activations lie on an $m$-dimensional subspace $\mathcal{S} \subset \mathbb{R}^d$ with orthonormal basis $U$, such that $f(x)=a = Uz$ for some latent $z \in \mathbb{R}^m$.
    \item \textbf{Linear Backbone:} The backbone classifier $g$ is linear prior to a softmax, i.e., $g(a) = \mathrm{softmax}(W_g a)$. 
    This reflects the standard post-hoc CBM setup of inserting the concept bottleneck at the penultimate layer, e.g., ~\cite{koh2020concept,ramaswamy2023overlooked,yuksekgonul2023posthoc}.
    \item \textbf{Random Projection:} The projection is defined as $\pi_\theta(a)=\sigma(Pa)$, where $P \in \mathbb{R}^{K \times d}$ is a fixed Gaussian matrix ($K \geq m$) and $\sigma$ is bijective (e.g., $\mathrm{tanh}$).
\end{enumerate}
Under these assumptions, we can construct a classifier $h$ that exactly recovers the original prediction $g(f(x))$:
\begin{equation} 
h(c) := g(a_{\mathrm{rec}}(c)) = \mathrm{softmax}(W_g \cdot [\underbrace{U (P U)^\dagger \sigma^{-1}(c)}_{\text{Reconstructed } a \in \mathcal{A}}]),
\end{equation}
where $a_{\mathrm{rec}}: \mathcal{C} \to \mathcal{S}$ reconstructs the activations $a=a_{\mathrm{rec}}(\pi_\theta(a))$ and $^\dagger$ denotes the Moore-Penrose pseudoinverse~\cite{penrose1956best}. 
We provide the full derivation in \Cref{apx:h-derivation}.

\bfparagraph{Consequences.}
Under the discussed assumptions, $h$ does not need to learn a decision based on semantic concepts. 
Instead, a degenerate solution is to reconstruct the original activations $a$ before applying $g$.
This aligns with \cite{srivastava2024vlg}, who demonstrated that random concept projections achieve zero expected error as the concept dimension $K$ approaches the activation dimension $d$.

Our analysis adds to this result in two aspects:
First, under the manifold hypothesis ($m \ll d$)~\cite{ansuini2019intrinsic,pope2021intrinsic}, the manifold JL lemma~\cite{baraniuk2009random} implies that $K$ need not approach $d$ to enable recovery.
Second, beyond bounding the expected error~\cite{srivastava2024vlg}, our constructive derivation under additional linearity assumption using \cite{penrose1956best} demonstrates that exact, sample-wise reconstruction may be possible.

Consequently, if $h$ is optimized rather than fixed, downstream accuracy becomes a poor proxy for concept faithfulness after \Cref{def:faithful}.
This necessitates a direct analysis of the learned projection $\pi_\theta$, which we formalize next.

\subsection{Faithfulness of the Learned Concept Projection}
\label{sec:method-concept-projection}

Assuming access to a training set $\mathbb{X}_{\mathrm{task}} = \{x_i, c_i\}_{i=1}^N$ for $\pi_\theta$ sampled i.i.d. from the task distribution $P_{\mathrm{task}}$, we minimize the expected risk $J_\mathrm{task}(\theta;\mathcal{L})$:
\begin{align}\label{eq:exp-risk}
   J_\mathrm{task}(\theta;\mathcal{L}) &= \mathbb{E}_{(x,c) \sim P_{\mathrm{task}}} \left[ \mathcal{L}(\pi_\theta(f(x)), \, c) \right] \\
   &\approx \frac{1}{N} \sum_{i=1}^N \mathcal{L}(\pi_\theta(f(x_i)), \, c_i). 
\end{align}
To interpret this optimization as a Maximum Likelihood Estimation (MLE), we assume that the projection $\pi_\theta$ defines the parameters of a conditional distribution $p_{\theta}(\cdot \mid f(x))$ over the concept space.
The empirical risk minimization becomes equivalent to MLE if the loss function $\mathcal{L}$ matches the negative log-likelihood of observing the true concept $c$ under this model distribution~\cite{Goodfellow.2016,Bishop.2006}:
\begin{equation}
    \mathcal{L}(\pi_\theta(f(x)), \, c) = - \log p_{\theta}(c \mid f(x)) + \text{const.}
\end{equation}
Consequently, the learned projection $\pi_\theta$ represents the optimal estimator for the concept scores distributed according to $P_{\mathrm{task}}(x,c)=P_\mathrm{task}(x)P_\mathrm{task}(c|x)$.
It minimizes \Cref{eq:exp-risk} and is, therefore, faithful according to \Cref{def:faithful}.

In practice, we identify two key problems that violate this theoretically optimal scenario:
First, the dataset used to learn the concept projection may differ from the task we are interested in solving, e.g., \cite{kim2018interpretability,yuksekgonul2023posthoc,schmalwasser2025fastcav}.
Intuitively, using an auxiliary concept dataset rather than a task-relevant $\mathbb{X}_\mathrm{task}$ leads to covariate shift, i.e., $P_\mathrm{aux}(x)\neq P_\mathrm{task}(x)$.
Second, to address this, prior work aims to assign task-relevant labels to in-domain training data, leading to surrogate labels $\tilde{c}$ with $P_\mathrm{task}(\tilde{c}|x)\neq P_\mathrm{task}(c|x)$, e.g., \cite{oikarinen2023label,srivastava2024vlg}.
Both scenarios are distributional mismatches (infracting \Cref{def:faithful}) while fitting $\pi_\theta$, which we will detail next.

\subsubsection{Covariate Shift of Concept Datasets.}
Training the projection $\pi_{\theta}$ requires access to a training set annotated with ground-truth concepts.
Because these are rarely available for the downstream task data $\mathbb{X}_{\mathrm{task}}$, we typically train $\pi_{\theta}$ by minimizing the empirical risk on a smaller auxiliary dataset $\mathbb{X}_{\mathrm{aux}}$ (e.g., Broden~\cite{bau2017network}) sampled from a distribution $P_{\mathrm{aux}}$~\cite{kim2018interpretability,yuksekgonul2023posthoc}.

However, this introduces the risk of distribution shift ($P_{\mathrm{task}} \neq P_{\mathrm{aux}}$).
Intuitively, images of a concept may appear visually different in $\mathbb{X}_\mathrm{aux}$ than in $\mathbb{X}_\mathrm{task}$, degrading the faithfulness of the learned concept projection after \Cref{def:faithful}.
In particular, the optimization objective in \Cref{eq:exp-risk} changes to minimizing a corresponding $J_\mathrm{aux}$.
To quantify this effect, we utilize an upper bound for the target generalization error from domain adaptation theory (Theorem 2 in \cite{BenDavid.2010}):
\begin{equation}
    \label{eq:cov-shift-error}
    J_{\mathrm{task}}(\theta;L_1) \leq J_{\mathrm{aux}}(\theta;L_1) + \frac{1}{2} \hat{d}_{\mathcal{H}\Delta\mathcal{H}}\!(\mathbb{X}_{\mathrm{task}},\!\mathbb{X}_{\mathrm{aux}}) + \Omega(\mathrm{dim}_{\mathrm{VC}},\!N,\!\delta) + \lambda_{\mathrm{ideal}}.
\end{equation}
Here, the expected $L_1$ (absolute) task risk $J_\mathrm{task}(\theta;L_1)$ is bounded by the expected auxiliary risk $J_\mathrm{aux}(\theta;L_1)$ plus the scaled empirical $\mathcal{H}\Delta\mathcal{H}$-divergence $\hat{d}_{\mathcal{H}\Delta\mathcal{H}}$~\cite{kifer2004hdivergence,BenDavid.2010} between the marginal distributions.
The term $\Omega$ accounts for finite sample estimation and depends on the VC dimension $\mathrm{dim}_{\mathrm{VC}}$, sample size $N$ (assuming $|\mathbb{X}_{\mathrm{aux}}| \approx |\mathbb{X}_{\mathrm{task}}|$), and the desired confidence level $\delta$.
$\lambda_{\mathrm{ideal}}$ represents the task's adaptability (the combined error of the optimal hypothesis on both distributions).

In concept probing, we assume this shift is primarily a \emph{Covariate Shift}~\cite{shimodaira2000covariateshift,sugiyama2012covariateshiftadaptation}, meaning the semantic definition of the concept $P_\mathrm{aux}(c|x)=P_\mathrm{task}(c|x)$ remains constant while the input marginals $P_\mathrm{aux}(x)\neq P_\mathrm{task}(x)$ change.
Under this assumption, there exists a single optimal decision rule shared across distributions, rendering $\lambda_{\mathrm{ideal}}$ negligible (full formal definitions and derivations in \Cref{app:covariate_shift}).

\bfparagraph{Consequences.}
When training on in-domain data is infeasible, we can estimate the faithfulness of $\pi_\theta$ after \Cref{def:faithful} on a given auxiliary dataset via this upper bound (\Cref{eq:cov-shift-error}).
Assuming a fixed $\pi_\theta$ architecture and equal dataset sizes (rendering $\Omega$ constant), and with $\lambda_{\mathrm{ideal}}$ negligible, the empirical divergence $\hat{d}_{\mathcal{H}\Delta\mathcal{H}}$ becomes the decisive metric for unfaithfulness.
Following~\cite{BenDavid.2010}, we approximate $\hat{d}_{\mathcal{H}\Delta\mathcal{H}}$ by training a domain discriminator to separate $\mathbb{X}_\mathrm{aux}$ and $\mathbb{X}_\mathrm{task}$.
Intuitively, if a simple classifier can distinguish the auxiliary and task data, we expect a larger absolute task error $J_\mathrm{task}(\theta;L_1)$, violating \Cref{def:faithful}.

\subsubsection{Faithfulness of Surrogate Label Functions.}
To avoid potential covariate shifts and the resulting generalization penalty created by training the projection $\pi_{\theta}$ on a separate auxiliary dataset, it would be ideal to train $\pi_{\theta}$ directly on the in-domain data $\mathbb{X}_{\mathrm{task}}$ sampled from $P_\mathrm{task}$ using an oracle concept function $c^*: \mathcal{X} \to \mathcal{C}$ with $\forall (x, c) \in \mathbb{X}_{\mathrm{task}}: c^*(x) = c$.
However, because these ground-truth annotations are commonly unavailable in post-hoc settings, we must rely on a surrogate labeling function $\tilde{c}: \mathcal{X} \to \mathcal{C}$ (e.g., derived from VLMs~\cite{oikarinen2023label,srivastava2024vlg,sampat2024help}; see \Cref{apx:vlm-surrogate-functions} for examples) to identify the concepts.

Consequently, the optimization objective shifts to minimizing the empirical risk with respect to the surrogate labels:
\begin{equation}
    \tilde{J}_\mathrm{task}(\theta;\mathcal{L}) = \mathbb{E}_{x \sim P_{\mathrm{task}}} \left[ \mathcal{L}\left(\pi_\theta(f(x)), \, \tilde{c}(x) \right) \right].
\end{equation}
The bottleneck's faithfulness thus depends on whether the parameters $\tilde{\theta}$ optimized for the surrogate $\tilde{c}$ also minimize the true risk with respect to $c^*$.

\bfparagraph{Theoretical Analysis.}
We analyze this condition by formulating $\pi_\theta$ as a \emph{Multivariate Generalized Linear Model} (GLM)~\cite{Fahrmeir.2001,McCullagh.1989}.
We assume the conditional distribution of the concepts belongs to the \emph{Multivariate Exponential Dispersion Family (EDF)}~\cite{McCullagh.1989,Jrgensen.1987}, which encompasses standard loss settings such as Mean Squared Error (Gaussian) and Cross-Entropy (Bernoulli/Multinomial).
In Appendix~\ref{apx:surrogate-faithfulness-derivation}, we derive the gradient of the \emph{true} objective $J^*_\mathrm{task}(\theta;\mathcal{L})$ at the \emph{surrogate} optimum $\tilde{\theta}$.
For canonical link functions, this gradient is proportional to the outer product ($\otimes$) of the surrogate error and backbone activations $f(x)$:
\begin{align}
\begin{split}
    \label{eq:true_grad_surrogate}
    \nabla \! J^*_\mathrm{task}(\tilde{\theta};\mathcal{L}) \! &= \! \mathbb{E}_{P_{\mathrm{task}}} \! \big[ \nabla \! \mathcal{L}(\pi_{\tilde{\theta}}(f(x)), c^*(x)) \big] \! \propto \! \mathbb{E}_{P_{\mathrm{task}}} \! \big[\! \underbrace{ ( \tilde{c}(x) \! - \! c^*(x) ) }_{\text{\tiny Surrogate Error $\delta(x)$}} \! \otimes \! f(x) \big].
\end{split}
\end{align}

For a faithful post-hoc CBM (i.e., $\tilde{\theta}$ is a global minimum of the objective, see \Cref{def:faithful}), this gradient must be zero.
This reveals two distinct mechanisms for faithfulness of $\pi_\theta$ trained to emulate the surrogate label function $\tilde{c}$:
\begin{enumerate}[label=(\arabic*)]
    \item \textbf{Surrogate Accuracy:} The surrogate labels have to be accurate, i.e., $\tilde{c}(x) \approx c^*(x)$, causing the surrogate error term $\delta(x)=\tilde{c}(x) - c^*(x)$ to vanish.
    \item \textbf{Error Orthogonality:} Further, \Cref{eq:true_grad_surrogate} implies that a model can remain faithful even with noisy surrogate labels if the surrogate error is \emph{orthogonal} to the activation space in expectation.
    This requires the errors in the surrogate labels to be uncorrelated with the activations $f(x)$.
\end{enumerate}

\begin{figure}[t]
\centering

\begin{subfigure}[b]{0.48\textwidth}
    \centering
    \resizebox{\linewidth}{!}{%
    \begin{tikzpicture}[
        scale=1.4, %
        vec/.style={-Stealth, thick},
        axis/.style={-Latex, black},
        ]
        
        \draw[axis] (-2.2,0) -- (2.2,0) node[below right, black] {Value for Sample $r$};
        \draw[axis] (0,-0.3) -- (0,1.7) node[above left, black] {Value for Sample $s$};
        
        \coordinate (O) at (0,0);
        \coordinate (Vf) at (-1.5, 0.8);  %
        \coordinate (Vd) at (0.8, 1.5);   %
        
        \draw[vec, red, line width=1.2pt] (O) -- (Vf) node[anchor=south east, text=red] {$\boldsymbol{\alpha}_{j}$};
        \draw[vec, blue, line width=1.2pt] (O) -- (Vd) node[anchor=west, text=blue] {$\mathbf{\Delta}_{k}$};
        
        \coordinate (P1) at ($(O)!0.25cm!(Vf)$);
        \coordinate (P2) at ($(O)!0.25cm!(Vd)$);
        \draw[black] (P1) -- ($(P1)+(P2)$) -- (P2);

        \node at (0.8, -0.2) [black] {Projection = $\mathbf{0}$};

    \end{tikzpicture}
    }
    \caption{Faithful Model (Orthogonal Case)}
    \label{fig:ortho_a}
\end{subfigure}
\hfill %
\begin{subfigure}[b]{0.48\textwidth}
    \centering
    \resizebox{\linewidth}{!}{%
    \begin{tikzpicture}[
        scale=1.4, %
        vec/.style={-Stealth, thick},
        axis/.style={-Latex, black},
        ]
        
        \draw[axis] (-2.2,0) -- (2.2,0) node[below right, black] {Value for Sample $r$};
        \draw[axis] (0,-0.3) -- (0,1.7) node[above left, black] {Value for Sample $s$};
        
        \coordinate (O) at (0,0);
        \coordinate (Vf) at (1.8, 0.6);         %
        \coordinate (Vd_non_ortho) at (1.0, 1.5); %
        
        \draw[vec, red, line width=1.2pt] (O) -- (Vf) node[anchor=north west, text=red] {$\boldsymbol{\alpha}_{j}$};
        \draw[vec, blue, dashed, line width=1.2pt] (O) -- (Vd_non_ortho) node[anchor=south, text=blue] {$\mathbf{\Delta}_{k}$};
        
        \coordinate (Proj) at ($(O)!(Vd_non_ortho)!(Vf)$);
        \draw[black, densely dotted, line width=0.8pt] (Vd_non_ortho) -- (Proj);
        \draw[vec, black, line width=1.5pt] (O) -- (Proj) node[anchor=north west, yshift=-4pt, xshift=-14pt, inner sep=2pt, black] {Projection $\neq \mathbf{0}$};

    \end{tikzpicture}
    }
    \caption{Unfaithful Model (Non-Orthogonal Case)}
    \label{fig:ortho_b}
\end{subfigure}

\caption{Visualization of the error orthogonality condition in a 2D sample space. 
The axes represent two samples $r$ and $s$ from the training set of $\pi_{\theta}$.
(a) A faithful model is learned if the surrogate label errors of these samples ($\mathbf{\Delta}_{k}$) are orthogonal to their activations ($\boldsymbol{\alpha}_{j}$), resulting in a zero projection. 
(b) Non-orthogonal errors have a non-zero projection onto the feature vector, resulting in a non-zero gradient for the true objective.
}
\label{fig:orthogonality_comparison}
\end{figure}
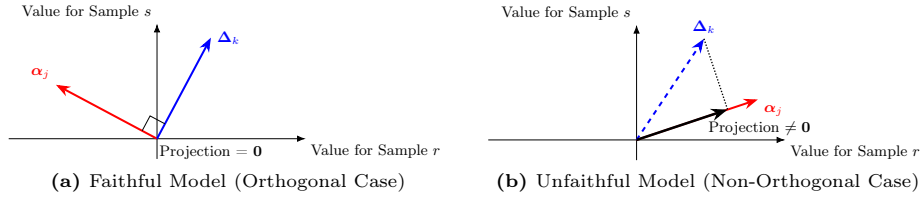

\bfparagraph{Geometric Interpretation.}
\Cref{fig:orthogonality_comparison} visualizes the orthogonality condition.
We consider $N$ training samples yielding activation vectors 
$\boldsymbol{\alpha}_{\!j} = [f_j(x_1)\!,\dots,\!f_j(x_N)]^\top$ 
for backbone neuron $j$, and surrogate discrepancy vectors 
$\mathbf{\Delta}_{\!k} = [\delta_k(x_1)\!,\dots,\!\delta_k(x_N)]^\top$ 
for concept $k$.
\Cref{eq:true_grad_surrogate} shows that faithfulness does not strictly require perfect labels ($\mathbf{\Delta}_{k} = \mathbf{0}$).
Instead, it is sufficient if the errors $\mathbf{\Delta}_{k}$ are orthogonal to the activations $\boldsymbol{\alpha}_j$ ($\langle \mathbf{\Delta}_{k}, \boldsymbol{\alpha}_{j} \rangle \approx 0$).
In other words, random, unsystematic annotation noise in the surrogate labels can cancel out during training.
However, \emph{systematic} errors (such as a VLM consistently predicting the concept ``Boat'' whenever it detects a ``Water'' texture, even if no boat is present), create non-orthogonal error components.
This results in a non-zero gradient for the true objective, causing the learned $\pi_{\tilde{\theta}}$ to diverge from the ground-truth semantics, violating \Cref{def:faithful}.

\bfparagraph{Consequences.}
We therefore propose two metrics to evaluate the faithfulness of surrogate-based post-hoc CBMs.
First, \emph{Surrogate Accuracy} directly measures the raw discrepancy $\delta(x)$.
Second, we evaluate the \emph{Alignment of Surrogate Errors and Activations} using the absolute Pearson correlation~\cite{pearson1896correlation} $\rho_{k,j} = \left| \mathrm{corr}(\mathbf{\Delta}_k, \boldsymbol{\alpha}_j) \right|$.
A correlation of $\rho \approx 0$ indicates orthogonal noise.
Conversely, if $|\rho|$ is large and significant (e.g., $p < 0.05$), the surrogate noise is systematically tied to the backbone features.
Thus, a post-hoc CBM is unfaithful only if surrogate labels are inaccurate \emph{and} those errors systematically correlate with backbone features.

\section{Experiments}

To validate our theoretical findings, we evaluate the faithfulness of concept projections $\pi_\theta$ trained using various post-hoc CBM methods.
First, we demonstrate that downstream performance is insufficient to assess concept bottleneck layers by studying random concept projections.
Next, we analyze the learned $\pi_\theta$ directly, leveraging a synthetic dataset with ground-truth concept information.
This allows us to measure the concept accuracy and the geometric alignment of the concept bottleneck approaches.
Using an optimal ground-truth classifier for the downstream task, we can align these results with downstream performance.
Finally, we evaluate our two theoretically identified sources of reduced faithfulness: concept-set covariate shift and systematic surrogate labeling errors.

\subsection{Experiment Setup}
\begin{figure}[t]
    \newcommand{\figwidth}{0.24} %

    \centering
    \begin{subfigure}[t]{\figwidth\linewidth}
        \centering
        \includegraphics[width=\linewidth]{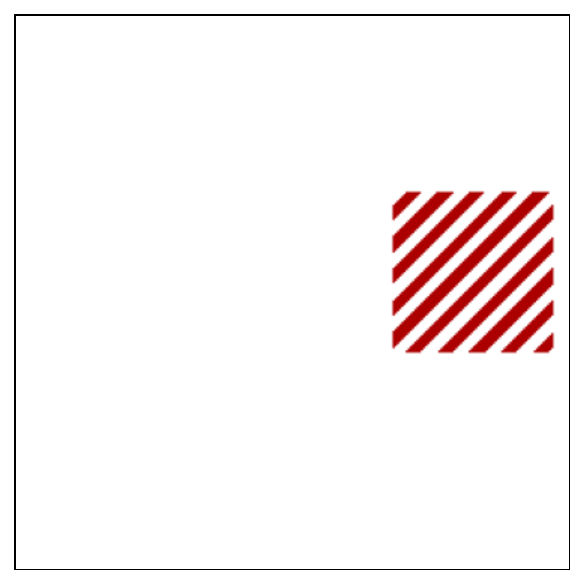}
        \caption{Example of $P_{\mathrm{task}}$}
        \label{fig:downstream_elements}
    \end{subfigure}
    \hfill
    \begin{subfigure}[t]{\figwidth\linewidth}
        \centering
        \includegraphics[width=\linewidth]{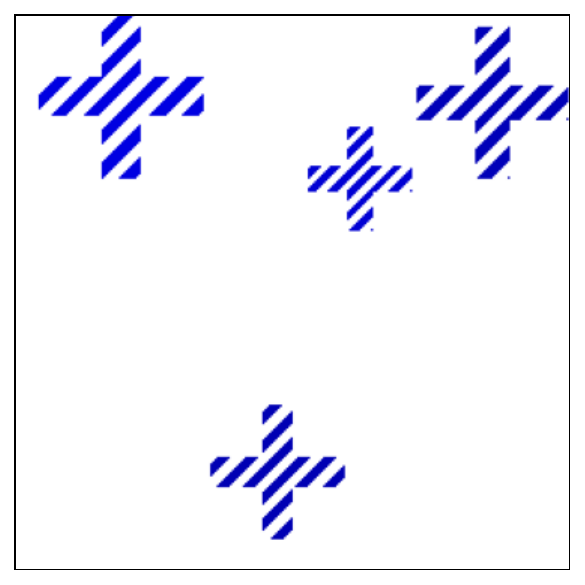}
        \caption{Example of $P_{\mathrm{near}}^{\forall}$}
        \label{fig:training_domain_only_concept_objects_elements}
    \end{subfigure}
    \hfill
    \begin{subfigure}[t]{\figwidth\linewidth}
        \centering
        \includegraphics[width=\linewidth]{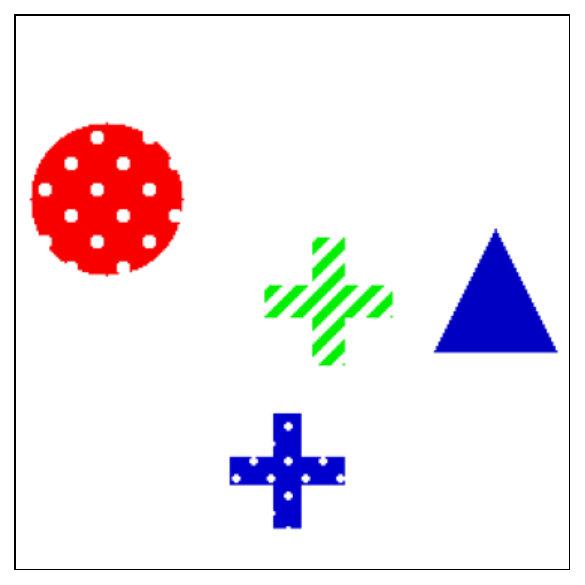}
        \caption{Example of $P_{\mathrm{near}}^{\exists}$}
        \label{fig:training_domain_objects_elements}
    \end{subfigure}
    \hfill
    \begin{subfigure}[t]{\figwidth\linewidth}
        \centering
        \includegraphics[width=\linewidth]{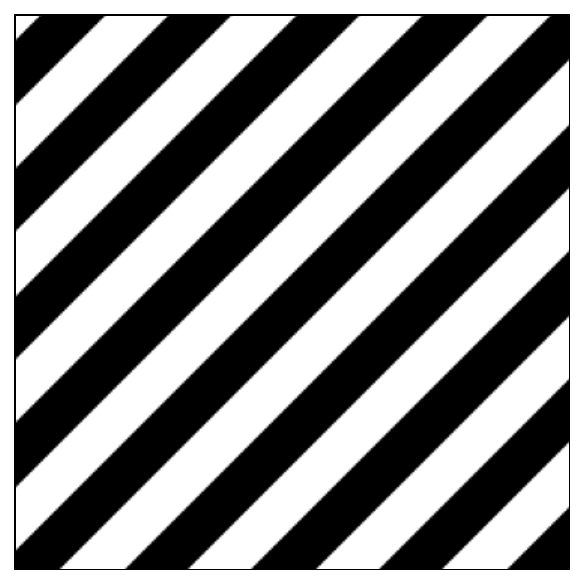}
        \caption{Example of $P_{\mathrm{OOD}}$}
        \label{fig:ood_elements}
    \end{subfigure}
    \caption{Samples of the concept ``striped'' for each visual set used.
    In \Cref{fig:downstream_elements}, there is exactly one object with the concept. \Cref{fig:training_domain_only_concept_objects_elements} shows a sample from the backbone domain where all objects share the concept. \Cref{fig:training_domain_objects_elements} shows a sample from the training domain and \Cref{fig:ood_elements} shows a complete out-of-domain example. 
    }
    \label{fig:visual_concepts}
\end{figure}

\paragraph{Datasets and Backbone Models.}
We validate our theoretical findings on standard real-world benchmarks: CUB-200~\cite{welinder2010cub} using a ResNet18 backbone \cite{he2016resnet,oleg2024imgclsmob} and CIFAR-10/100~\cite{krizhevsky2009cifar} using CLIP ResNet-50 (RN50) \cite{deng2009imagenet} as backbone.
Additionally, we consider the synthetic \emph{Elements} dataset \cite{nicolson2025explaining}, which includes ground-truth concept annotations.

\emph{Elements} samples consist of objects defined by three latent attributes: shape, color, and texture (e.g., ``red striped square'').
To create a realistic shift between backbone pretraining and downstream application, we train the backbone $f$ to identify these objects in a multi-object classification task following \cite{nicolson2025explaining}, whereas the downstream post-hoc CBM task involves classifying images containing only a \emph{single} object.
In all experiments, the backbone $f$ remains frozen.

\paragraph{Probing Datasets for Concept Learning.}
For standard PCBMs \cite{yuksekgonul2023posthoc}, we train the concept projection $\pi_\theta$ on different auxiliary distributions, while keeping the semantic concept definitions fixed (visualized in \Cref{fig:visual_concepts}):
\begin{itemize}
    \item \textbf{In-Domain:}
    We train $\pi_\theta$ using samples from the exact downstream distribution $P_{\mathrm{task}}$.
    For \emph{Elements}, this corresponds to single-object images exhibiting the specific concept (e.g., all images containing ``red'').
    \item \textbf{Near-Domain:} Concepts are sampled from a distribution that is distinct from, but visually similar to $P_{\mathrm{task}}$.
    For \emph{Elements}, we consider $P_{\mathrm{near}}^{\forall}$ with multiple objects per image that all share concepts including the target, and $P_{\mathrm{near}}^{\exists}$, where at least one object exhibits the target concept.
    \item \textbf{Out-of-Distribution:}
    Concepts are learned from a visually distinct distribution $P_{\mathrm{OOD}}$ (e.g., abstract patterns) that share the concept label but differ structurally from the downstream task.
    This represents the domain gap common in post-hoc analysis (e.g., datasets like Broden~\cite{bau2017network}).
\end{itemize}

\paragraph{Post-Hoc Concept Bottleneck Models (post-hoc CBMs).}
In our experiments, we train post-hoc CBMs with two strategies: training $\pi_{\theta}$ on an auxiliary concept dataset and using a VLM to generate surrogate concept labels.
Specifically, we utilize a VLM (CLIP~\cite{radford2021learning} and DINOv3~\cite{simeoni2025dinov3}, each with a ViT-B/16~\cite{dosovitskiy2021transformer}) as the backbone $f$ and learn $\pi_{\theta}$ using the text embedding for each concept \cite{yuksekgonul2023posthoc}.
Additionally, we consider Label-Free CBMs (LFCBMs)~\cite{oikarinen2023label}, where surrogate concept labels for $P_{\mathrm{task}}$ are generated by computing the similarity between the CLIP-representation~\cite{radford2021learning} of the input image and the text embeddings for the considered concepts.
Further, we evaluate Vision-Language-Guided CBMs (VLG-CBMs)~\cite{srivastava2024vlg}, which use the open-vocabulary object detector Grounding DINO~\cite{liu2024grounding} to generate binary surrogate concept labels for $P_{\mathrm{task}}$ for each concept based on the predicted presence of concepts in the image \cite{srivastava2024vlg}.
For all post-hoc CBM variants, the bottleneck size $K$ is determined by the number of concepts derived from the concept source (we consider Broden~\cite{bau2017network}, ConceptNet 5.5~\cite{speer2017conceptnet}, and GPT-3~\cite{brown2020language}).

\subsection{The Illusion of Classifier Accuracy}
\label{sec:random-exp}
\begin{figure}[!t]
    \centering
    \includegraphics[width=\linewidth]{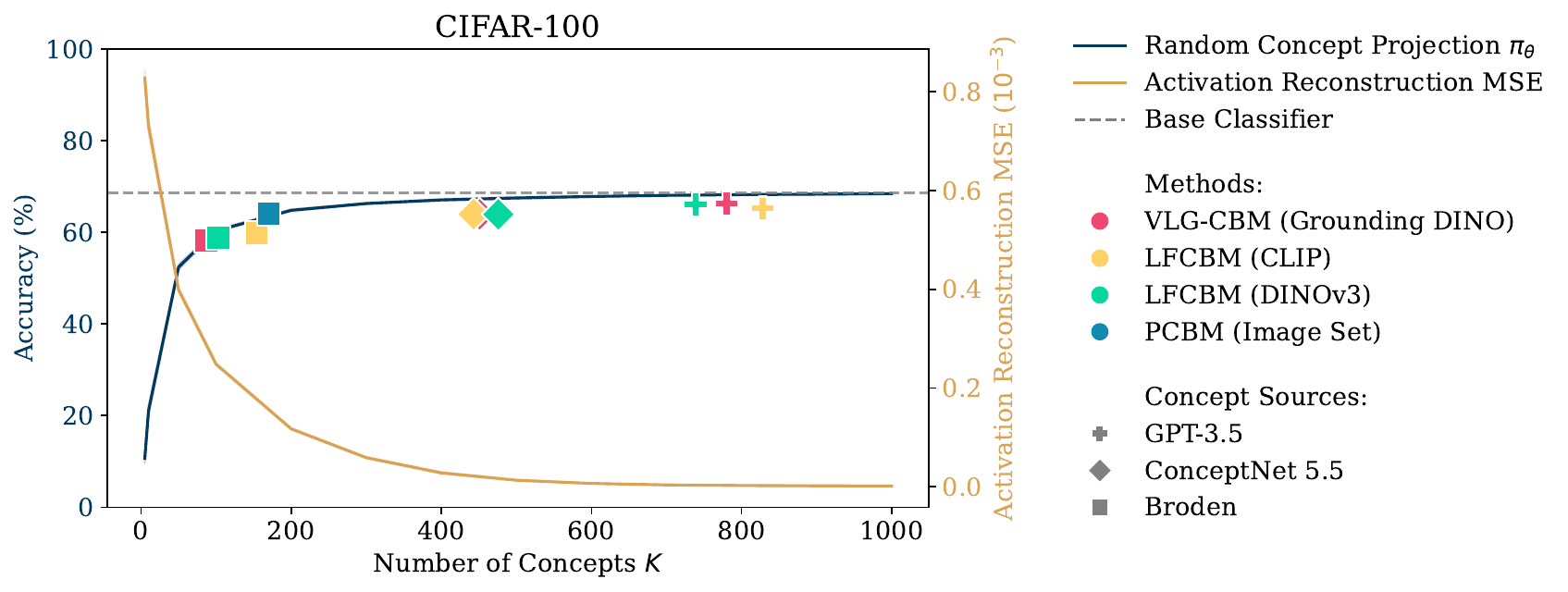}
    \caption{High downstream accuracy does not imply faithfulness.
    The dashed line shows the classifier's performance without a bottleneck layer.
    The downstream classification accuracy of a post-hoc CBM increases with the number of random concept vectors.
    Additionally, various post-hoc CBMs that use meaningful concept sources only marginally outperform random concepts.
    Hence, downstream performance does not indicate faithful concept representations, which aligns with our theoretical findings in \Cref{sec:method-downstream}.
    }
    \label{fig:random_directions_accuracy}
\end{figure}

\bfparagraph{Setup.}
To validate that downstream accuracy is an insufficient metric for concept bottleneck quality (\Cref{sec:method-downstream}), we follow a similar setup to \cite{midavaine2024reproducibility} and build post-hoc CBMs using frozen, standard-normal random projections $\pi_\theta$ of increasing dimension $K$.
For comparison, we train post-hoc CBMs using dominant paradigms from the literature: 
(1) Auxiliary concept datasets \cite{yuksekgonul2023posthoc}, e.g., Broden \cite{bau2017network}, and 
(2) VLM surrogate labeling \cite{oikarinen2023label,srivastava2024vlg} using either the same concepts as in (1) or external sources (ConceptNet 5.5 \cite{speer2017conceptnet} or GPT 3.5 \cite{brown2020language}).
In all cases, the downstream classifier $h$ is trained after freezing $\pi_\theta$ (see \Cref{apx:random-concepts} for additional details).

\bfparagraph{Results.}
\Cref{fig:random_directions_accuracy} demonstrates that for CIFAR-100 \cite{krizhevsky2009cifar} (other datasets in \Cref{apx:random-concept-add-results}), increasing the bottleneck dimension $K$ of random projections decreases activation reconstruction error and closes the downstream accuracy gap to a fully supervised, bottleneck-free backbone.
Notably, the post-hoc CBMs~\cite{yuksekgonul2023posthoc,oikarinen2023label,srivastava2024vlg}  trained on common concept sources~\cite{bau2017network,speer2017conceptnet,brown2020language} achieve similar performance to random projections of equal size $K$.
This confirms that high downstream accuracy is insufficient evidence for a meaningful bottleneck, requiring a direct concept projection analysis instead.

\subsection{Quantifying Misalignment and Concept Projection Performance}

\paragraph{Setup.}
Having shown that downstream accuracy is deceptive, we now directly measure the faithfulness of $\pi_{\theta}$.
We evaluate post-hoc CBMs trained on four auxiliary distributions ($P_{\mathrm{task}}$, $P_{\mathrm{near}}^{\forall}$, $P_{\mathrm{near}}^{\exists}$, $P_{\mathrm{OOD}}$; \Cref{fig:visual_concepts}), alongside VLM-surrogate models: LFCBM~\cite{oikarinen2023label} (using CLIP~\cite{radford2021learning} and DINOv3~\cite{simeoni2025dinov3}) and VLG-CBM~\cite{srivastava2024vlg} (using Grounding DINO~\cite{liu2024grounding}).
We calculate the concept accuracy of $\pi_{\theta}$ on both its training ($P_{\mathrm{aux}}$) and downstream ($P_{\mathrm{task}}$) distributions, and geometric faithfulness via the cosine similarity between learned concept vectors $v_k$ and target ground-truth directions $v_k^*$: $\mathrm{sim}_\sphericalangle = (v_k \cdot v_k^*) / (\|v_k\| \|v_k^*\|)$ (\Cref{apx:alignment-cav}).
To further determine if $h$ masks identified unfaithfulness, we compare its accuracy against an oracle classifier $h^*$ that uses the optimal rule for mapping ground-truth concepts to classes (cf. independent bottleneck strategy~\cite{koh2020concept}).
Though related to information leakage (see \Cref{sec:related-work}), faithfulness is complementary. Unfaithfulness can occur without leakage and vice versa (analyzed in \Cref{apx:alternative-faithfulness-metrics-results}).

\begin{table}[t]
\centering
\small
\setlength{\tabcolsep}{4pt}
\setlength{\defaultaddspace}{3pt} %
\caption{Faithfulness vs. Performance. Comparison of post-hoc CBM variants with respect to the ground truth concept labels and direction on the Elements dataset \cite{nicolson2025explaining}. OOD data degrades faithfulness significantly more than accuracy suggests. Each metric is averaged over the $K$ concepts. To demonstrate faithfulness under error orthogonality, we also train $\pi_\theta$ on ground-truth concept labels while adding 25\% random noise.}
\resizebox{\textwidth}{!}{
\begin{tabular}{llccccc}
\toprule
& & \multicolumn{3}{c}{Concept Projection $\pi_{\theta}$} & \multicolumn{2}{c}{Downstream Classifier} \\
\cmidrule(lr){3-5} \cmidrule(lr){6-7}
Method & \multirow{-2}{*}{\begin{tabular}[b]{@{}l@{}} Dataset/\\Label \end{tabular}} & $\mathrm{sim}_\sphericalangle$ $[\uparrow]$ & Acc. ${P_{\mathrm{aux}}}$ $[\uparrow]$ & Acc. ${P_{\mathrm{task}}}$ $[\uparrow]$ & Acc. $h$ $[\uparrow]$ & Acc. $h^*$ $[\uparrow]$ \\

\midrule
\multicolumn{7}{l}{\textit{(1) Training on Auxiliary Distributions (Oracle Labels)}} \\
\addlinespace
\multirow{4}{*}{PCBM~\cite{yuksekgonul2023posthoc}} & $P_{\mathrm{task}}$ & 0.96{\scriptsize $\pm$0.01} & 1.00{\scriptsize $\pm$0.00} & 1.00{\scriptsize $\pm$0.00} & 1.00{\scriptsize $\pm$0.00} & 1.00{\scriptsize $\pm$0.00} \\
 & $P_{\mathrm{near}}^{\forall}$ & 0.97{\scriptsize $\pm$0.01} & 1.00{\scriptsize $\pm$0.00} & 0.70{\scriptsize $\pm$0.02} & 1.00{\scriptsize $\pm$0.00} & 0.99{\scriptsize $\pm$0.01} \\
 & $P_{\mathrm{near}}^{\exists}$ & 0.57{\scriptsize $\pm$0.02} & 0.82{\scriptsize $\pm$0.02} & 0.89{\scriptsize $\pm$0.01} & 1.00{\scriptsize $\pm$0.00} & 0.85{\scriptsize $\pm$0.04} \\
 & $P_{\mathrm{OOD}}$ & 0.47{\scriptsize $\pm$0.01} & 0.98{\scriptsize $\pm$0.01} & 0.77{\scriptsize $\pm$0.01} & 1.00{\scriptsize $\pm$0.00} & 0.21{\scriptsize $\pm$0.05} \\

\midrule
\multicolumn{7}{l}{\textit{(2) Surrogate Concept Labels (Training Dist.: $P_{\mathrm{task}}$)}} \\
\addlinespace
\addlinespace
\multirow{3}{*}{LFCBM~\cite{oikarinen2023label}} & {25\% Label Noise} & 0.02{\scriptsize $\pm$0.03} & 0.75{\scriptsize $\pm$0.01} & 0.98{\scriptsize $\pm$0.02} & 1.00{\scriptsize $\pm$0.00} & 0.98{\scriptsize $\pm$0.03} \\
& CLIP~\cite{radford2021learning} & 0.05{\scriptsize $\pm$0.02} & 0.79{\scriptsize $\pm$0.01} & 0.56{\scriptsize $\pm$0.02} & 0.99{\scriptsize $\pm$0.00} & 0.54{\scriptsize $\pm$0.04} \\
 & DINOv3~\cite{simeoni2025dinov3} & 0.05{\scriptsize $\pm$0.01} & 0.85{\scriptsize $\pm$0.01} & 0.58{\scriptsize $\pm$0.02} & 1.00{\scriptsize $\pm$0.00} & 0.78{\scriptsize $\pm$0.05} \\
\addlinespace
VLG-CBM~\cite{srivastava2024vlg} & G--DINO~\cite{liu2024grounding} & 0.02{\scriptsize $\pm$0.02} & 0.74{\scriptsize $\pm$0.01} & 0.85{\scriptsize $\pm$0.01} & 1.00{\scriptsize $\pm$0.01} & 0.04{\scriptsize $\pm$0.01} \\
\bottomrule
\end{tabular}
}
\label{tab:elements_faithfulness}
\end{table}

\paragraph{Results.}
\Cref{tab:elements_faithfulness} summarizes the results for the Elements dataset \cite{nicolson2025explaining}, where we have access to ground-truth concept labels.
We split our analysis into the two primary post-hoc CBM setups: (1) training on auxiliary datasets \cite{yuksekgonul2023posthoc} and (2) using in-domain surrogate concept labels \cite{oikarinen2023label,srivastava2024vlg}.

For (1), all models achieve high accuracy on $P_{\mathrm{aux}}$, but $P_\mathrm{OOD}$ in particular yields unfaithful target representations, as indicated by the concept accuracy and the geometric alignment $\mathrm{sim}_\sphericalangle$.
Crucially, the learned $h$ masks these failures with high downstream accuracy.
In contrast, the oracle $h^*$ demonstrates the unfaithfulness.
In particular, the model trained on $P_\mathrm{OOD}$ drops to $21\%$ accuracy, consistent with the expected impact of a severe shift in concept representation.

For (2), LFCBMs \cite{oikarinen2023label} with noisy ground-truth labels (25\% random noise) predict target concepts accurately.
This matches our expectation from \Cref{sec:method-concept-projection} that unsystematic label errors average out during the training of $\pi_\theta$.
In contrast, LFCBMs~\cite{oikarinen2023label} trained with CLIP~\cite{radford2021learning} and DINOv3~\cite{simeoni2025dinov3} surrogate labels, achieve poor performance in the target distribution, though DINOv3 improves marginally upon CLIP.
These observations are confirmed by the learned classifiers $h$, which show near-perfect accuracy for all settings.
However, the oracle $h^*$ can again confirm the underlying unfaithfulness.
VLG-CBMs~\cite{srivastava2024vlg} using Grounding DINO~\cite{liu2024grounding} exacerbate this: $h$ perfectly solves the task, but $h^*$ beats random guessing ($\nicefrac{1}{36}\approx 0.028$) only marginally.
To explain the representation failures demonstrated in this experiment, we next investigate our theoretically derived sources of unfaithful concept projections $\pi_\theta$ (\Cref{sec:method-concept-projection}).

\subsection{Measuring the Impact of Covariate Shift}
\label{sec:covariate-shift-exp}

\begin{figure}[t]
    \newcommand{\imgwidth}{1.0} %
    \centering
    \begin{subfigure}[t]{0.48\textwidth}
        \centering
        \includegraphics[width=\imgwidth\linewidth]{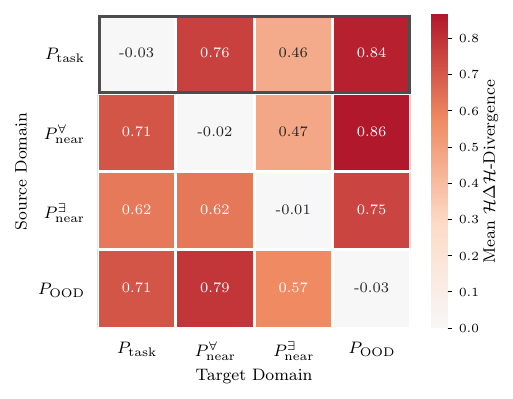}
        \caption{Average Pairwise $\mathcal{H}\Delta\mathcal{H}$-Divergence Matrix}
        \label{fig:pairwise-h-divergences}
    \end{subfigure}
    \hfill
    \begin{subfigure}[t]{0.48\textwidth}
        \centering
        \includegraphics[width=\imgwidth\linewidth]{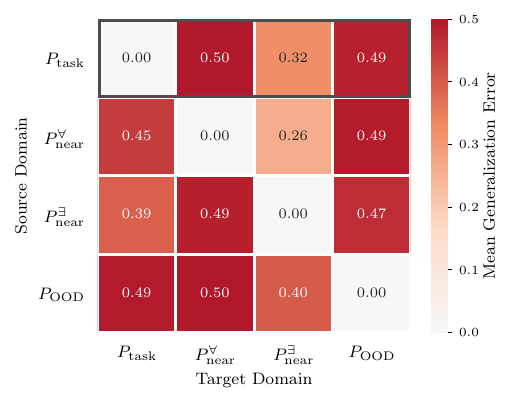}
        \caption{Pairwise Generalization Errors}
        \label{fig:pairwise-generalization-errors}
    \end{subfigure}
    \caption{Visualization of the estimated $\mathcal{H}\Delta\mathcal{H}$-divergence as a proxy for the upper bound of the task domain error.
    We pairwise compare the four probing datasets for the synthetic Elements downstream task \cite{nicolson2025explaining}, highlighting the actual downstream task in our setup.
    Note that the colormaps in (a) and (b) use different scales.
    We find that stronger covariate shifts lead to increased unfaithfulness, whereas lower divergence (i.e., in-domain concept sets) yields improved $\pi_\theta$.
    }
    \label{fig:H-divergence}
\end{figure}

\paragraph{Setup.}
To evaluate the impact of covariate shifts when training $\pi_{\theta}$ on an auxiliary dataset, we estimate the $\hat{d}_{\mathcal{H}\Delta\mathcal{H}}$-divergence~\cite{kifer2004hdivergence,BenDavid.2010} between source and target domains, which serves as an upper bound to the generalization error (\Cref{sec:method-concept-projection}).
Following~\cite{BenDavid.2010}, we approximate it by training a modified Huber loss discriminator~\cite{Zhang.2004} on the backbone activations to separate the distributions, computing $\hat{d}_{\mathcal{H}\Delta\mathcal{H}} \approx 2(1 - 2\epsilon)$ based on the discriminator's classification error $\epsilon$.
We then compare these estimates against the actual generalization errors of $\pi_\theta$.

\paragraph{Results.}
As shown in \Cref{fig:H-divergence} (with additional details and results in \Cref{apx:covariate-shift-results}), empirical divergences align closely with actual $\pi_\theta$ generalization errors. 
The out-of-distribution domain $P_\mathrm{OOD}$ consistently yields the highest divergences and generalization errors.
Conversely, matched source and target domains (on the diagonal) yield near-zero empirical divergences, confirming the high PCBM \cite{yuksekgonul2023posthoc} performance observed in \Cref{tab:elements_faithfulness} (slightly negative values occur if the discriminator's accuracy falls marginally below 0.5).
Among mismatched domains, $P_\mathrm{near}^\exists$ and $P_\mathrm{near}^\forall$ show the lowest divergence, reflecting the latter being a subdistribution of the former.
Overall, we find a strong correlation ($>0.9$) between the estimated $\hat{d}_{\mathcal{H}\Delta\mathcal{H}}$ and actual generalization errors, demonstrating that $\hat{d}_{\mathcal{H}\Delta\mathcal{H}}$ is an effective proxy for the unfaithfulness introduced by a covariate shift of the auxiliary concept set.
This is highly practical because it can be estimated entirely without ground-truth concept labels for the downstream task.

\subsection{Measuring Systematic Surrogate Label Errors}
\label{sec:surrogate-label-errors-exp}

\paragraph{Setup.}
Following \Cref{sec:method-concept-projection}, we analyze the faithfulness of post-hoc CBMs trained with surrogate labels based on the surrogate error $\mathbf{\Delta}_k$ for concept $k$ using two metrics: (1) error magnitude (mean absolute error), and (2) error orthogonality (absolute Pearson correlation~\cite{pearson1896correlation} $|\rho_{k,j}|$ between surrogate label errors and backbone activations $\boldsymbol{\alpha}_j$).
Because calculating $\mathbf{\Delta}_k$ requires ground truth concepts, we evaluate on the Elements dataset~\cite{nicolson2025explaining}.
We compare surrogate labels generated with Grounding DINO~\cite{liu2024grounding,srivastava2024vlg} against a baseline with unsystematic label noise obtained by randomly flipping a portion of the ground-truth concept labels (see \Cref{apx:systematic-surrogate-errors-results} for CLIP~\cite{radford2021learning}, DINOv3~\cite{simeoni2025dinov3}, and CUB-200~\cite{welinder2010cub}).

\begin{figure}[!tb]
    \newcommand{\imgwidth}{1.0} %

    \centering
    \begin{subfigure}[t]{0.47\textwidth}
        \centering
        \includegraphics[width=\imgwidth\linewidth]{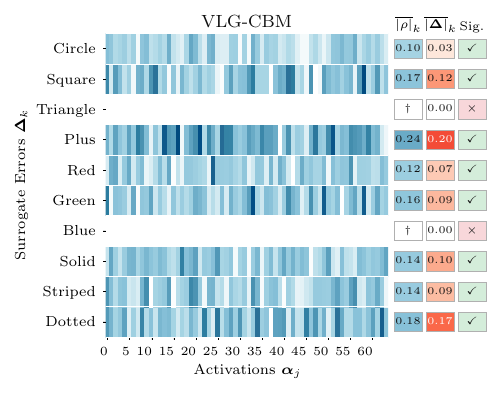}
        \caption{Systematic Surrogate Error Introduced by Grounding DINO}
        \label{fig:sys_noise_vlg}
    \end{subfigure}
    \hfill
    \begin{subfigure}[t]{0.52\textwidth}
        \centering
        \includegraphics[width=\imgwidth\linewidth]{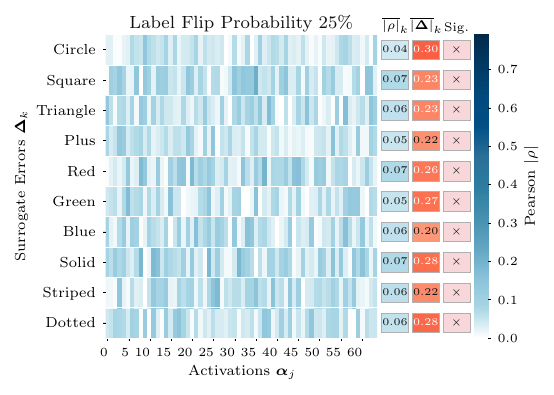}
        \caption{Unsystematic (Random) Surrogate Error}
        \label{fig:random_noise}
    \end{subfigure}
    \caption{
        Systematic vs. Unsystematic Surrogate Errors.
        Unfaithful projections occur when surrogate errors are both large ($\overline{|\mathbf{\Delta}|}_k$) and highly correlated ($\overline{|\rho|}_k$) with activations $\boldsymbol{\alpha}_j$.
        (a) Surrogates from Grounding DINO~\cite{liu2024grounding} exhibit systematic, correlated errors, causing unfaithful projections.
        (b) Random unsystematic label noise produces no correlation.
        If the VLM perfectly predicts a concept ($\overline{|\mathbf{\Delta}|}_k = 0$), correlation is undefined ($\dagger$).
    }
        
    \label{fig:pearson_correlation_heatmap}
\end{figure}

\paragraph{Results.}
\Cref{fig:sys_noise_vlg} shows these correlations alongside per-concept summary statistics $\overline{|\rho|}_k$ and $\overline{|\mathbf{\Delta}|}_k$ (calculated as the absolute average over per-activation scores, using a Holm-Bonferroni correction~\cite{holm1979correction} at $p=0.05$ to determine significance).
As established in \Cref{eq:true_grad_surrogate}, optimizing $\pi_\theta$ yields unfaithful representations when both values are large.
Unlike the random-noise baseline, where the correlations effectively vanish, VLM surrogate errors are clearly systematic and correlated with specific activations.
Qualitatively, we find that Grounding DINO~\cite{liu2024grounding} struggles to positively detect these concepts, which we attribute to the domain shift between its natural-image pretraining and our synthetic setup (geometric shapes on white backgrounds).
Because the VLM label noise is non-random, training on it can cause $\pi_\theta$ to learn unfaithful projections (see (2) in \Cref{tab:elements_faithfulness}).

While faithfulness is preserved for concepts that are labeled perfectly by the VLM (e.g., ``Triangle'' and ``Blue'' for Grounding DINO~\cite{liu2024grounding}), achieving purely random, uncorrelated surrogate errors may be difficult in many real-world scenarios.
Since we identify systematic surrogate label errors as a source of unfaithfulness, correlation can be used to rank imperfect surrogate labeling functions by the severity of this failure mode.
For $P_{\mathrm{task}}$, the average $\rho$ across concepts is lowest for Grounding DINO~\cite{liu2024grounding} ($0.142$), followed by CLIP~\cite{radford2021learning} ($0.152$) and DINOv3~\cite{simeoni2025dinov3} ($0.155$), making Grounding DINO the preferred surrogate labeling function under this criterion.

\section{Conclusions}
While post-hoc CBMs promise transparent decision-making, current evaluations often conflate predictive performance with conceptual alignment.
As our theoretical analysis based on \cite{baraniuk2009random} shows, even non-semantic random projections achieve competitive accuracy.
Consequently, downstream task performance is an uninformative metric for concept bottleneck quality, requiring the direct inspection of the concept projection $\pi_\theta$.
Doing so reveals two primary sources of unfaithfulness in standard post-hoc CBM training.
First, relying on auxiliary datasets may introduce a covariate shift that can invalidate learned concepts in the target domain.
Based on \cite{BenDavid.2010}, we provide an upper bound on the generalization error and a practical metric to measure it.
Second, we demonstrate that systematic label errors introduced by VLM surrogate supervision lead to unfaithfulness.
By formalizing these issues and validating them across real-world and synthetic benchmarks, we establish an evaluation framework for post-hoc CBM unfaithfulness.

These insights open several avenues for future research.
Our formalization of distribution shifts suggests the need for explicit domain adaptation techniques within the post-hoc CBM framework to align feature spaces, rather than just labels.
Furthermore, the presence of systematic VLM label noise requires the development of de-biasing mechanisms that disentangle correlated errors during surrogate supervision.
Ultimately, we provide the theoretical and practical tools necessary to investigate learned concept projections.

\bibliographystyle{splncs04}
\bibliography{main}

\newpage
\appendix

\crefalias{section}{appendix}
\crefalias{subsection}{appendix}

\onecolumn
\section{Additional Theoretical Details}
\subsection{Johnson-Lindenstrauss for Smooth Manifolds}
\label{apx:jl-lemma}

\begin{appendixblock}

While neural activation spaces are high-dimensional ($\mathcal{A}\subset \mathbb{R}^d$), valid activations empirically concentrate on a manifold $\mathcal{M}$ of much lower intrinsic dimension $m\ll d$ \cite{ansuini2019intrinsic, pope2021intrinsic}.
If this manifold hypothesis of deep learning holds, then the Johnson-Lindenstrauss (JL) lemma for smooth manifolds \cite{baraniuk2009random} governs the behavior under random projections $\pi_\theta: \mathcal{A}\rightarrow\mathcal{C}\subset \mathbb{R}^K$.
Additionally, while \cite{makonnen2025measuring} discusses this for finite datasets (the standard JL lemma \cite{johnson1984extensions}), we note that \cite{baraniuk2009random} implies that it holds for the complete activation manifold.

Next, we follow \cite{baraniuk2009random} to make the manifold JL lemma explicit for our use case of concept embeddings.
First, for the standard JL lemma \cite{johnson1984extensions}, the number of random projections needed mainly depends on the number of points and the original dimension ($d$ in our case).
For the manifold version, it depends on the manifold's properties \cite{baraniuk2009random}.
Let $\mathcal{M}\subset \mathbb{R}^d$ be a compact $m$-dimensional Riemannian submanifold with a volume $V$,  a condition number $\nicefrac{1}{\tau}$, and the geodesic covering regularity $R$ after \cite{baraniuk2009random}.
According to the Manifold Johnson-Lindenstrauss lemma (Thm. 3.1 in \cite{baraniuk2009random}), a random projection into $K$ dimensions preserves pairwise Euclidean distances on $\mathcal{M}$ with distortion $0<\epsilon <1$ with probability $1-p$ if
\begin{equation}\label{eq:mjl}
     K = \mathcal{O}\left(\frac{m \log(dVR\tau^{-1}\epsilon^{-1})\log(p^{-1})}{\epsilon^2}\right).
\end{equation}
Crucially, $K$ scales linearly with the intrinsic dimension $m$ but logarithmically with the activation space dimension $d$. 
Thus, if $K \ll d$ sufficiently exceeds $m$, the geometry of $\mathcal{A}$ is preserved in $\mathcal{C}$.
There is also a logarithmic dependence on the volume $V$, the geodesic covering regularity $R$, and the condition number $\nicefrac{1}{\tau}$.
Specifically, the condition number bounds the curvature of the manifold \cite{baraniuk2009random}.
Thus, highly curved manifolds (large $\nicefrac{1}{\tau}$) need a larger number of random projections.
We found this also in our concrete derivation for a linear subspace \Cref{apx:h-derivation}, where the intrinsic dimension $m$ is enough.

In any case, a sufficiently expressive nonlinear classifier $h: \mathcal{C}\rightarrow \mathcal{Y}$, e.g., an MLP \cite{hornik1989multilayer}, can approximate $g\circ\pi_\theta^{-1}$ and solve the downstream task with high performance.
This insight aligns with performance gains observed using non-linear classifiers on concept bottlenecks \cite{yuksekgonul2023posthoc,zarlenga2022concept}.

Crucially, our discussion here is based on the manifold hypothesis.
While there is empirical evidence that this applies in common cases, e.g., for realistic datasets \cite{ansuini2019intrinsic,pope2021intrinsic}, we want to stress that settings where it does not hold may be possible.
Note that in such cases \cite{srivastava2024vlg} proves that the expected error under random projections goes towards zero for the embedding dimension $K$ approaching the ambient dimensions $d$.

\end{appendixblock}

\newpage
\subsection{Derivation of a Sufficient Classifier Under Linear Subspace Assumption}
\label{apx:h-derivation}

\begin{appendixblock}

As mentioned in our main paper, for sufficiently expressive $h$, it is possible to achieve high performance even for random concept projections $\pi_\theta$. 
Our argument is based on inverting the original activations $a = f(x)$.
Following the assumptions stated in our main paper, we derive $h$ as follows:

Let $\sigma^{-1}(c) = Pa$ be the pre-activation concepts before the application of the bijective activation function $\sigma$.
Following Assumption (1), we have $\sigma^{-1}(c) = PUz$.
Let $M=PU\in \mathbb{R}^{K\times m}$ be the matrix product of the random projection matrix $P$ and the orthonormal basis $U$ of the linear activation subspace $\mathcal{S}$.
Since $U$ has full rank and $P$ is a standard Gaussian matrix, $M$ is full column rank with probability 1 (almost surely) if $K\geq m$ (Assumption (2) and Assumption(3)).
This is implied by the two-sided bounds of the singular values (Thm.~4.6.1 in \cite{vershynin2018high}) for such matrices.

Consequently, $M$ is left-invertible. 
The latent vector $z$ can be recovered exactly using the Moore-Penrose pseudoinverse $M^\dagger = (M^\top M)^{-1}M^\top$ \cite{penrose1956best}:
\begin{equation}
z = M^\dagger \sigma^{-1}(c) = (PU)^\dagger \sigma^{-1}(c).
\end{equation}
That $\sigma^{-1}$ exists follows from the bijectivity (Assumption (3)).
The original activations $a$ are reconstructed as
\begin{equation}
\hat{a} = U z = U (PU)^\dagger \sigma^{-1}(c)
\end{equation}
and we define $a_{\mathrm{rec}}: \mathcal{C} \to \mathcal{S}$ with $a_{\mathrm{rec}}(c):= U (PU)^\dagger \sigma^{-1}(c)$ as the function reconstructing the original activations for a given $c \in \mathcal{C}$.
Finally, we construct the sufficient classifier $h$ to mimic the original $g$
\begin{align}
\nonumber h(c) &:= g(a_{\mathrm{rec}}(c))\\ %
&= \mathrm{softmax}\left( W_g \cdot \left[ U (P U)^\dagger \sigma^{-1}(c) \right] \right).
\end{align}

\end{appendixblock}

\subsection{Formalization of the Covariate Shift Assumption}
\label{app:covariate_shift}

\begin{appendixblock}

In \Cref{sec:method-concept-projection}, we utilize the domain adaptation bound from \cite{BenDavid.2010} to bound the expected $L_1$ (absolute) task risk after training on auxiliary concept sets.
Here, we provide formal definitions for the task adaptability term and the covariate shift assumption.

The final term of the bound, $\lambda_{\mathrm{ideal}}$, represents the combined error of the optimal hypothesis that performs best on both the auxiliary and task distributions simultaneously:
\begin{equation}
    \lambda_{\mathrm{ideal}} = \min_{\theta}  \Big( \underbrace{\mathbb{E}_{x \sim P_{\mathrm{aux}}} [|\pi_{\theta}(f(x)) - c^*_{\mathrm{aux}}(x)|]}_{\text{Error vs Probing Set Rule}} + \underbrace{\mathbb{E}_{x \sim P_{\mathrm{task}}} [|\pi_{\theta}(f(x)) - c^*_{\mathrm{task}}(x)|]}_{\text{Error vs Downstream Task Rule}} \Big),
\end{equation}
where $c^*_{\mathrm{aux}}$ and $c^*_{\mathrm{task}}$ denote the underlying ground-truth concept labeling functions for the respective domains, $P_\mathrm{aux}$ and $P_\mathrm{task}$.

In the context of concept probing, we assume the distribution shift is primarily a \emph{Covariate Shift}~\cite{shimodaira2000covariateshift,sugiyama2012covariateshiftadaptation}.
This implies that while the probability of observing specific input samples differs between the domains, the semantic definition of the concepts remains stable (i.e., the conditional probability of the concept given the input is constant):
\begin{align}
    &\forall x, c \in \mathcal{X} \times \mathcal{C}: P_{\mathrm{task}} (c|x) = P_{\mathrm{aux}}(c|x), \text{~and~} \exists x \in \mathcal{X}: P_{\mathrm{task}} (x) \neq P_{\mathrm{aux}}(x).
\end{align}
Under this assumption, there exists a single optimal decision rule $c^*$ shared across distributions.
Consequently, $\lambda_{\mathrm{ideal}}$ simplifies to the combined error of the best shared projection:
\begin{equation}
    \lambda_{\mathrm{ideal}} = \min_{\theta}  \Big( \underbrace{\mathbb{E}_{x \sim P_{\mathrm{aux}}} [|\pi_{\theta}(f(x)) - c^*(x)|]}_{\text{Error vs Shared Rule}} + \underbrace{\mathbb{E}_{x \sim P_{\mathrm{task}}} [|\pi_{\theta}(f(x)) - c^*(x)|]}_{\text{Error vs Shared Rule}} \Big).
\end{equation}
As argued in the main text, this allows us to treat $\lambda_{\mathrm{ideal}}$ as negligible and isolate the empirical $\mathcal{H}\Delta\mathcal{H}$-divergence as our proxy for the unfaithfulness of $\pi_\theta$ trained on the auxiliary concept dataset $\mathbb{X}_\mathrm{aux}$.

\end{appendixblock}

\subsection{Vision-Language-Based Surrogate Labeling Functions}
\label{apx:vlm-surrogate-functions}

\begin{appendixblock}

As noted in our main paper, to minimize domain shift when learning the concept projection $\pi_\theta$, it is advisable to directly train on in-domain data with respect to the downstream task.
However, often concept annotations in sufficient quantity are unavailable directly.
Hence, surrogate labeling functions based on vision language models (VLMs) are used in practice, e.g., \cite{oikarinen2023label,srivastava2024vlg,sampat2024help}, to approximate the true labeling function $c^*$.
Note that in any case, sets of concepts as text descriptions may be derived from Large Language Models (LLMs) specifically for the downstream classes \cite{oikarinen2023label,srivastava2024vlg,sampat2024help}.
Here, we describe two current frameworks for annotating such concepts for in-domain data.

\bfparagraph{Surrogate Labels From Vision-Language Embeddings \protect{\cite{oikarinen2023label}}.}
A promising strategy for generating surrogate concept labels is to leverage the zero-shot capabilities of vision-language encoders (VLEs) \cite{oikarinen2023clip,oikarinen2023label}.
Let $e_V:\mathcal{X}\to\mathbb{R}^d$ and $e_T:\mathcal{T}\to\mathbb{R}^d$ denote the vision and text encoders of a pre-trained VLM (e.g., CLIP~\cite{radford2021learning}), sharing a joint embedding space.
Given a set of concept prompts $\mathbb{T}=\{t_k\}_{k=1}^K$, we define the surrogate labeling function $\tilde{c}_{\mathrm{VLE}} : \mathcal{X} \to \mathcal{C}$ using a similarity metric $\mathrm{sim}: \mathbb{R}^d \times \mathbb{R}^d \to \mathbb{R}$.
For an input $x$, the $K$ concept scores are computed as:
\begin{equation}
    \tilde{c}_{\mathrm{VLE}}(x) \;=\;
\big(\mathrm{sim}(e_V(x), e_T(t_k))\big)_{k=1}^K \in \mathbb{R}^K.
\end{equation}
Common choices for the similarity metric include standard cosine similarity, $s_{\cos}(u, v) = \frac{\langle u, v \rangle}{\|u\|\|v\|}$, or sharpened variants designed to suppress noise from low-similarity such as the cubed cosine similarity $s_{\cos^3}(u, v) = (s_{\cos}(\bar{u}^3, \bar{v}^3))$, where $\bar{u}$ and $\bar{v}$ are standardized to mean 0 and variance 1~\cite{oikarinen2023label}.
Consequently, a post-hoc CBM with VLE-based surrogate concept label generation optimizes:
\begin{equation}
    \tilde{J}_\mathrm{task}(\theta;\mathcal{L}) = \mathbb{E}_{x \sim P_{\mathrm{task}}} \left[ \mathcal{L}\left(\pi_\theta(f(x)), \, \tilde{c}_{\mathrm{VLE}}(x)\right) \right],
\end{equation}
thereby aligning model activations with VLM-derived surrogate concepts.
In practice, we use CLIP \cite{radford2021learning} and DINOv3 \cite{simeoni2025dinov3} encoders in our experiments.

\bfparagraph{Surrogate Labels From Grounded Object Detection \protect{\cite{srivastava2024vlg}}.}
Alternatively, surrogate concepts can be generated by spatially grounding language concepts using open-vocabulary object detectors (e.g., Grounding-DINO~\cite{liu2024grounding}).
Following the VLG-CBM \cite{srivastava2024vlg} approach, an LLM is employed to generate a set of candidate concept prompts $\mathbb{T}=\{t_k\}_{k=1}^K$ relevant to the classes in $\mathcal{Y}$.
To obtain the surrogate labels, a grounded object detector $D_{\mathrm{obj}}$ takes an image $x \in \mathcal{X}$ and the prompt set $\mathbb{T}$ as input, and returns a set of detected bounding boxes with associated confidence scores.
Let the output detections be $\mathcal{B}(x) = \{(b_m, \sigma_m, k_m)\}_{m=1}^M$, where $b_m$ represents the box coordinates, $\sigma_m \in [0,1]$ is the confidence score, and $k_m \in \{1, \dots, K\}$ is the index of the matched concept $t_{k_m}$.
Then, the grounded surrogate labeling function $\tilde{c}_{\mathrm{VLG}} : \mathcal{X} \to \{0,1\}^K$ is based on thresholding the detection scores:
\begin{equation}
    \tilde{c}_{\mathrm{VLG}}(x) \;=\; (\mathds{1}\left[ \exists m : k_m = k \land \sigma_m > \tau \right])_{k=1}^K,
\end{equation}
where $\tau$ is a pre-defined confidence threshold and $\mathds{1}[\cdot]$ is the indicator function.
Unlike the continuous scores discussed above for $\tilde{c}_{\mathrm{VLE}}$, this method yields a binary concept vector.
The post-hoc CBM is then trained using these binary labels, typically treating the concept alignment as a multi-label classification task, optimizing:
\begin{equation}
    \tilde{J}_\mathrm{task}(\theta;\mathcal{L}) = \mathbb{E}_{x \sim P_{\mathrm{task}}} \left[ \mathcal{L}\left(\pi_\theta(f(x)), \, \tilde{c}_{\mathrm{VLG}}(x)\right) \right].
\end{equation}

\bfparagraph{Surrogate Labels From Visual Question Answering \protect{\cite{sampat2024help}}.}
A direct alternative to bounding-box detection is to prompt advanced vision-language models, e.g., \cite{li2022blip,liu2024llavanext,comanici2025gemini,Qwen3-VL}, and phrasing surrogate labeling as a Visual Question Answering (VQA) problem \cite{sampat2024help}.
Unlike embedding-based methods that rely on latent space similarity, this strategy treats concept extraction as a Boolean verification task in language space.
Let $\Phi$ be a generative VLM that maps an image $x \in \mathcal{X}$ and a text prompt $p$ to a probability distribution over a vocabulary $\mathcal{V}$.
For each concept $t_k \in \mathbb{T}$, a concept query prompt $p_k$ is constructed, such as ``Is the concept '$t_k$' present in this image? Answer Yes or No.''.
The surrogate labeling function $\tilde{c}_{\mathrm{VQA}} : \mathcal{X} \to [0,1]^K$ is derived by computing the probability of the affirmative token $t_{\mathrm{yes}}$ (e.g., ``Yes'') given the image and prompt:
\begin{equation}
    \tilde{c}_{\mathrm{VQA}}(x) \;=\; (P_\Phi(t_{\mathrm{yes}} \mid x, p_k))_{k=1}^K.
\end{equation}
Depending on the implementation, this score may be used directly as a soft concept probability or is thresholded into a binary concept vector to optimize $\pi_\theta$:
\begin{equation}
    \tilde{J}_\mathrm{task}(\theta;\mathcal{L}) = \mathbb{E}_{x \sim P_{\mathrm{task}}} \left[ \mathcal{L}\left(\pi_\theta(f(x)), \, \tilde{c}_{\mathrm{VQA}}(x)\right) \right].
\end{equation}

\end{appendixblock}

\newpage
\subsection{Conditions for Surrogate Label Faithfulness}
\label{apx:surrogate-faithfulness-derivation}

\begin{appendixblock}

\begin{figure*}[t!]
    \centering
    \tikzset{
        c0s0/.style={circle, draw=blue!80!black, fill=blue!30, thick, minimum size=7pt, inner sep=0pt},
        c1s1/.style={rectangle, draw=red!80!black, fill=red!30, thick, minimum size=7pt, inner sep=0pt, sharp corners},
        c0s1/.style={circle, draw=blue!80!black, fill=red!30, thick, minimum size=7pt, inner sep=0pt},
        c1s0/.style={rectangle, draw=red!80!black, fill=blue!30, thick, minimum size=7pt, inner sep=0pt, sharp corners},
        bg0/.style={fill=blue!5},
        bg1/.style={fill=red!5}
    }

    \begin{adjustbox}{width=0.55\textwidth}
    \begin{tikzpicture}
        \node[draw=black!50, fill=white, rounded corners, inner sep=8pt, align=center] {
            \begin{tabular}{llcll}
                 \tikz[baseline=-0.8ex]{\node[c0s0] {};} \small True 0, Surr. 0 (Correct) &&&
                 \tikz[baseline=-0.8ex]{\node[c1s0] {};} \small True 1, Surr. 0 (\textbf{Error}) \\
                 \tikz[baseline=-0.8ex]{\node[c1s1] {};} \small True 1, Surr. 1 (Correct) &&&
                 \tikz[baseline=-0.8ex]{\node[c0s1] {};} \small True 0, Surr. 1 (\textbf{Error}) \\
                 \multicolumn{4}{l}{\tikz[baseline=-0.6ex]{\draw[thick, black] (0,0) -- (0.5,0);} \small Learned Boundary $\pi_\Theta$ \quad \tikz[baseline=-0.6ex]{\draw[thick, dotted, gray] (0,0) -- (0.5,0);} \small Ideal Boundary}
            \end{tabular}
        };
    \end{tikzpicture}
    \end{adjustbox}
    
    \vspace{0.5em} %

    \begin{subfigure}[b]{0.32\textwidth}
        \centering
        \begin{adjustbox}{width=\textwidth}
        \begin{tikzpicture}[scale=0.85, every node/.style={scale=0.85}]
            \fill[bg0] (-2.5, -2.5) rectangle (0, 2.5);
            \fill[bg1] (0, -2.5) rectangle (2.5, 2.5);
    
            \draw[->, gray] (-2.5,0) -- (2.5,0) node[right, black] {$f_1(x)$};
            \draw[->, gray] (0,-2.5) -- (0,2.5) node[above, black] {$f_2(x)$};
    
            \draw[thick, black] (0, -2.5) -- (0, 2.5);
    
            \node[c0s0] at (-1.5, 1.5) {};
            \node[c0s0] at (-0.5, 1.8) {};
            \node[c0s0] at (-1.2, 0.5) {};
            \node[c0s0] at (-0.6, 0.2) {};
            \node[c0s0] at (-1.8, -0.6) {};
            \node[c0s0] at (-0.4, -1.2) {};
            \node[c0s0] at (-1.5, -1.7) {};
            
            \node[c1s1] at (1.5, 1.5) {};
            \node[c1s1] at (0.5, 1.8) {};
            \node[c1s1] at (0.8, 2.2) {};
            \node[c1s1] at (1.2, 0.5) {};
            \node[c1s1] at (0.6, 0.2) {};
            \node[c1s1] at (1.8, -0.6) {};
            \node[c1s1] at (0.4, -1.2) {};
            \node[c1s1] at (1.5, -1.7) {};
        \end{tikzpicture}
        \end{adjustbox}
        \caption{True Concept \\Labels}
        \label{fig:surrogate_true}
    \end{subfigure}\hfill
    \begin{subfigure}[b]{0.32\textwidth}
        \centering
        \begin{adjustbox}{width=\textwidth}
        \begin{tikzpicture}[scale=0.85, every node/.style={scale=0.85}]
            \fill[bg0] (-2.5, -1.25) -- (2.5, 1.25) -- (2.5, 2.5) -- (-2.5, 2.5) -- cycle;
            \fill[bg1] (-2.5, -2.5) -- (-2.5, -1.25) -- (2.5, 1.25) -- (2.5, -2.5) -- cycle;
    
            \draw[->, gray] (-2.5,0) -- (2.5,0) node[right, black] {$f_1(x)$};
            \draw[->, gray] (0,-2.5) -- (0,2.5) node[above, black] {$f_2(x)$};
    
            \draw[thick, dotted, gray] (0, -2.5) -- (0, 2.5);
    
            \draw[thick, black] (-2.5, -1.25) -- (2.5, 1.25);
            
            \draw[-Stealth, thick, black!70] (0.1, 1.07) -- (1.9, 1.07) node[midway, below=1pt, font=\scriptsize] {Shift};
    
            \node[c0s0] at (-1.5, 1.5) {};
            \node[c0s0] at (-0.5, 1.8) {};
            \node[c0s0] at (-1.2, 0.5) {};
            \node[c0s0] at (-0.6, 0.2) {};
            \node[c0s0] at (-1.8, -0.6) {};
            \node[c0s0] at (-0.4, -1.2) {}; 
            \node[c0s0] at (-1.5, -1.7) {};
    
            \node[c1s0] at (1.5, 1.5) {}; %
            \node[c1s0] at (0.5, 1.8) {}; %
            \node[c1s0] at (0.8, 2.2) {}; %
            
            \node[c1s1] at (1.2, 0.5) {};
            \node[c1s1] at (0.6, 0.2) {};
            \node[c1s1] at (1.8, -0.6) {};
            \node[c1s1] at (0.4, -1.2) {};
            \node[c1s1] at (1.5, -1.7) {};
            
            \draw[thick, dashed, black!60] (1.07, 1.825) ellipse (0.9 and 0.65);
        \end{tikzpicture}
        \end{adjustbox}
        \caption{Systematic Concept \\Errors}
        \label{fig:surrogate_systematic}
    \end{subfigure}\hfill
    \begin{subfigure}[b]{0.32\textwidth}
        \centering
        \begin{adjustbox}{width=\textwidth}
        \begin{tikzpicture}[scale=0.85, every node/.style={scale=0.85}]
            \fill[bg0] (-2.5, -2.5) rectangle (0, 2.5);
            \fill[bg1] (0, -2.5) rectangle (2.5, 2.5);
    
            \draw[->, gray] (-2.5,0) -- (2.5,0) node[right, black] {$f_1(x)$};
            \draw[->, gray] (0,-2.5) -- (0,2.5) node[above, black] {$f_2(x)$};
    
            \draw[thick, black] (0, -2.5) -- (0, 2.5);
    
            \node[c0s1] (err1) at (-1.5, 1.5) {}; %
            \node[c0s0] at (-0.5, 1.8) {};
            \node[c0s0] at (-1.2, 0.5) {};
            \node[c0s0] at (-0.6, 0.2) {};
            \node[c0s0] at (-1.8, -0.6) {};
            \node[c0s1] (err2) at (-0.4, -1.2) {}; %
            \node[c0s0] at (-1.5, -1.7) {};
    
            \node[c1s1] at (1.5, 1.5) {};
            \node[c1s0] (err3) at (0.5, 1.8) {}; %
            \node[c1s1] at (0.8, 2.2) {};
            \node[c1s1] at (1.2, 0.5) {};
            \node[c1s1] at (0.6, 0.2) {};
            \node[c1s0] (err4) at (1.8, -0.6) {}; %
            \node[c1s1] at (0.4, -1.2) {};
            \node[c1s1] at (1.5, -1.7) {};
            
        \end{tikzpicture}
        \end{adjustbox}
        \caption{Random Concept\\Errors}
        \label{fig:surrogate_random}
    \end{subfigure}

    \caption{
    Geometric intuition for surrogate faithfulness based on a concept encoded as binary classification (shapes denote true concept labels according to the ground-truth concept labeling function $c^*$; colored borders denote surrogate labels generated with $\tilde{c}$).
    (a) The ideal boundary learned from true labels.
    (b) If the surrogate labeling function generates errors that are systematically correlated with specific features, the learned boundary differs from the ideal boundary, leading to unfaithfulness.
    (c) If errors in the surrogate labels are random (specifically, if $\mathbb{E}[\delta(x) \otimes f(x)] \approx 0$, meaning they are orthogonal to the backbone features in expectation), the ideal boundary is recovered. This allows us to reach a faithful $\pi_{\Theta}$ despite imperfect surrogate labels.
    }
    \label{fig:surrogate_theory_intuition}
\end{figure*}
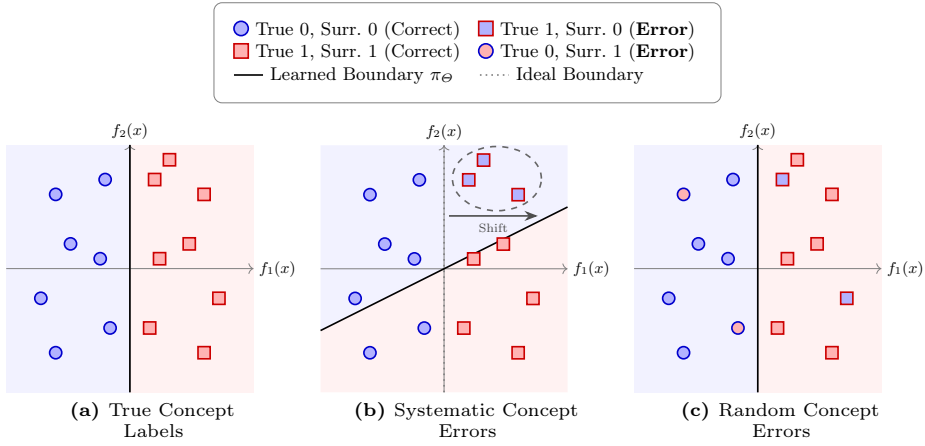

In the post-hoc CBM framework, the concept projection $\pi_\theta: \mathcal{A} \to \mathcal{C}$ maps the $d$-dimensional backbone activations $f(x) \in \mathcal{A} \subseteq \mathbb{R}^d$ to a $K$-dimensional concept space $\mathcal{C} \subseteq \mathbb{R}^K$ (\Cref{sec:method}).
Because a ground-truth labeling function $c^*$ is typically unavailable in post-hoc settings, this projection is often trained using a surrogate labeling function $\tilde{c}$ (\Cref{sec:method-concept-projection}), commonly derived from the zero-shot capabilities of VLMs (\Cref{apx:vlm-surrogate-functions}).
Consequently, the post-hoc CBM's faithfulness depends on whether training $\pi_\theta$ on the surrogate labels from $\tilde{c}$ results in a valid solution with respect to the true labels from $c^*$.
This is important to ensure the practical benefits of post-hoc CBMs, e.g., facilitating expert interventions \cite{yuksekgonul2023posthoc}.
If the concept predictions are not aligned with the human semantics encoded in $c^*$, interventions can be problematic \cite{zarlenga2022concept}, which is difficult to verify in latent concept bottle layers \cite{penzel2026locally}.

In the following, we show under which conditions we can learn a faithful concept projection $\pi_\theta$ based on $\tilde{c}$ (\Cref{fig:surrogate_theory_intuition}), i.e., minimizing $J_\mathrm{task}(\theta;\mathcal{L})$, \cref{def:faithful}.
We begin by formulating $\pi_\theta$ within the framework of \emph{Multivariate Generalized Linear Models} (GLM)~\cite{Fahrmeir.2001,McCullagh.1989}.
The GLM assumption captures the standard post-hoc CBM architecture, in which the concept projection $\pi_{\theta}$ is commonly represented by a linear layer. 
This restricts the parameters $\theta$ to a learnable matrix $\Theta \in \mathbb{R}^{K \times d}$ (the linear probe weights).
Consequently, the concept projection takes the form $\pi_\Theta(f(x))$.
Our analysis proceeds in five steps: First, we present the distributional (1) and structural (2) assumptions under which we model $\pi_{\theta}$ as a GLM.
Then, we derive the gradients for the log-likelihood for the true concept labels (3) and the surrogate objective (4), before finally using the gradient of the true objective (5) to derive the exact conditions for surrogate faithfulness after \cref{def:faithful}.

\bfparagraph{1. Distributional Assumption.}
Because concepts in post-hoc CBMs can be modeled in various ways, such as binary indicators (e.g., presence of ``wings'') or continuous similarity scores (e.g., similarity with a VLM prompt), we need a general framework to model the concept distribution.
The \emph{Multivariate Exponential Dispersion Family} (EDF)~\cite{McCullagh.1989,Jrgensen.1987} provides this generalization:
It is a broad family of probability distributions that generalizes almost all relevant distributions used in practice (including Normal, Binomial, Poisson, and Gamma distributions), offering a unified mathematical treatment for the conditional expectation $\mu$.
We assume the conditional distribution of the concept vector $c \in \mathcal{C}$ given an input $f(x) \in \mathcal{A}$ follows an EDF defined by a natural parameter vector $\eta \in \mathbb{R}^K$~\cite{McCullagh.1989,Jrgensen.1987}:
\begin{equation}
    p(c \mid \eta, \phi) = \exp\left( \frac{\langle c, \eta \rangle - A(\eta)}{s(\phi)} + k(c, \phi) \right),
\end{equation}
where:
\begin{itemize}
    \item $\langle \cdot, \cdot \rangle$ denotes the standard dot product in $\mathbb{R}^K$.
    \item $c \in \mathcal{C}$ is the observed concept vector.
    \item $\eta \in \mathbb{R}^K$ is the natural parameter vector determining the distribution's mean.
    \item $A: \mathbb{R}^K \to \mathbb{R}$ is the strictly convex \emph{cumulant function} (log-partition function).
    \item $\phi > 0$ is the scalar \emph{dispersion parameter} scaling the variance.
    \item $s(\phi)$ is the \emph{dispersion function}, a strictly positive scaling function (typically $s(\phi)=\phi$). %
    \item $k(c, \phi)$ ensures that the resulting distribution is normalized. Since it is independent of $\eta$, it vanishes upon differentiation. %
\end{itemize}
Here, only $\eta$ and $\phi$ are determined by $\pi_\Theta$, while all remaining parameters are fixed by the general distribution.
This framework covers discrete and continuous concept label distributions, such as binary concepts following a Bernoulli distribution ($\phi=1$) and continuous concepts following a Gaussian distribution ($\phi=\sigma^2$ for variance $\sigma^2$). 
A fundamental property of the EDF that we will leverage is that the expected concept vector $\mu = \mathbb{E}[c \mid \eta, \phi]$ is directly given by the gradient of the cumulant function~\cite{McCullagh.1989}: %
\begin{equation}
    \label{eq:edf_mean}
    \mu = \nabla_{\eta} A(\eta).
\end{equation}

\bfparagraph{2. Structural Assumption.}
To formally model the concept projection, we follow the standard GLM framework~\cite{Fahrmeir.2001, McCullagh.1989}, which connects the backbone activations to the expected concept values $\mu$ through a linear predictor $\zeta$ and an invertible link function $\psi$.
In the context of post-hoc CBMs, the linear predictor corresponds to the learnable linear layer applied to the backbone features.
We define this predictor as a function $\zeta: \mathcal{A} \to \mathbb{R}^K$, parameterized by the weight matrix $\Theta \in \mathbb{R}^{K \times d}$:
\begin{equation}
    \zeta(f(x)) = \Theta f(x).
\end{equation}
Because the linear output is unbounded while concept expectations may be constrained (e.g., $\mu \in [0,1]^K$), the link function $\psi: \mathcal{C} \to \mathbb{R}^K$ is used to enforce the relationship $\psi(\mu) \stackrel{!}{=} \zeta(f(x))$.
In neural network terms, the inverse link function $\psi^{-1}: \mathbb{R}^K \to \mathcal{C}$ corresponds to the activation function (e.g., sigmoid or softmax) that transforms the linear logits into the final prediction:
\begin{equation}
    \psi(\mu) = \zeta(f(x)) \Longleftrightarrow \mu = \psi^{-1}(\zeta(f(x))) = \psi^{-1}(\Theta f(x)).
\end{equation}
Specifically, we assume that $\psi$ is the \emph{canonical link function}, which equates the mean $\mu$ to the natural parameter $\eta$ such that $\psi(\mu) := \eta$~\cite{McCullagh.1989}: 
\begin{equation}
    \eta = \psi(\mu) = \zeta(f(x)) = \Theta f(x).
\end{equation}
This assumption covers standard choices for the combinations of distributional assumptions and activation functions (such as pairing a Bernoulli distribution with logit link) and simplifies the gradient derivations.
By combining this structural assumption with the EDF property from \Cref{eq:edf_mean}, the model's concept prediction $\hat{c}_\Theta(x)$ is given by the gradient of the cumulant function at the linear predictor:
\begin{equation}
    \hat{c}_\Theta(x) = \mu = \nabla_{\eta} A(\eta) = \nabla_{\Theta f(x)} A(\Theta f(x)).
\end{equation}

\bfparagraph{3. Gradient of the Log-Likelihood.}
Our optimization objective is to minimize the Negative Log-Likelihood (NLL) on the training data.
For a single observation $(f(x), c) \in \mathcal{A} \times \mathcal{C}$, the loss function $\mathcal{L}$ is derived from the EDF density:
\begin{align}
    \mathcal{L}(\pi_\Theta(f(x)), c) &= - \log p(c \mid \eta, \phi) \\ %
    &= - \log \exp\left( \frac{\langle c, \eta \rangle - A(\eta)}{s(\phi)} + k(c, \phi) \right) \\
    &= - \left( \frac{\langle c, \eta \rangle - A(\eta)}{s(\phi)} + k(c, \phi) \right) \\
    &\;\propto\; - \frac{1}{s(\phi)} \left( \langle c, \eta \rangle - A(\eta) \right) ,
\end{align}
where we have omitted terms independent of $\Theta$.
Substituting the canonical link $\eta = \Theta f(x)$, we differentiate with respect to the weight matrix $\Theta$.
Based on \Cref{eq:edf_mean}, we define the concept prediction as $\hat{c}_\Theta(x) = \mu = \nabla_{\eta} A(\eta)$.
Applying the chain rule for matrix calculus and leveraging the identity $\nabla_\Theta \langle u, \Theta v \rangle = u v^\top$, we obtain:
\begin{align}
    \nabla_\Theta \mathcal{L}(\pi_\Theta(f(x)), c) &= - \frac{1}{s(\phi)} \nabla_\Theta \left( \langle c, \eta \rangle - A(\eta) \right) \\
    &= - \frac{1}{s(\phi)} \left( \nabla_\Theta \left( c^\top \Theta f(x) \right) - \nabla_\Theta A(\Theta f(x)) \right) \\
    &= - \frac{1}{s(\phi)} ( \nabla_\Theta \left( c^\top \Theta f(x) \right) - \underbrace{\nabla_{\eta} A(\eta)}_{\hat{c}_\Theta(x)} f(x)^\top ) \\
    &= - \frac{1}{s(\phi)} \left( c f(x)^\top - \hat{c}_\Theta(x) f(x)^\top \right).
\end{align}
Rearranging the terms yields the final gradient form:
\begin{equation}
    \label{eq:gradient_mnef}
    \nabla_\Theta \mathcal{L}(\pi_\Theta(f(x)), c) = \frac{1}{s(\phi)} \underbrace{\left( \hat{c}_\Theta(x) - c \right)}_{\text{Residual}} \otimes \underbrace{f(x)}_{\text{Input}},
\end{equation}
where $\otimes$ denotes the outer product.
That is, for a distribution in the Multivariate EDF with a canonical link, the gradient is the outer product of the concept residual and the backbone activations, scaled by the inverse dispersion.

\bfparagraph{4. Surrogate Optimality Condition.}
We assume the projection parameters $\Theta$ are learned using a surrogate labeling function $\tilde{c}: \mathcal{X} \to \mathcal{C}$ (see \Cref{apx:vlm-surrogate-functions} for examples) rather than the inaccessible ground truth.
Thus, the surrogate objective $\tilde{J}_\mathrm{task}(\Theta;\mathcal{L})$ is the expected risk over the task distribution $P_{\mathrm{task}}$, calculated using these surrogate labels:
\begin{equation}
    \label{eq:surrogate-cost-function}
    \tilde{J}_\mathrm{task}(\Theta;\mathcal{L}) = \mathbb{E}_{x \sim P_{\mathrm{task}}} \left[ \mathcal{L}(\pi_\Theta(f(x)), \tilde{c}(x)) \right].
\end{equation}
Because we consider a GLM with a canonical link function, the negative log-likelihood is convex with respect to the linear parameters $\Theta$ (provided the cumulant function $A$ is convex, which holds by definition for our EDF)~\cite{McCullagh.1989}.
Consequently, the optimization problem is convex, and any stationary point corresponds to a global minimum.
We assume the optimization procedure reaches such a stationary point (e.g., via unregularized empirical risk minimization), yielding learned parameters $\tilde{\Theta} \in \mathbb{R}^{K \times d}$ that satisfy the first-order condition $\nabla \tilde{J}_\mathrm{task}(\tilde{\Theta};\mathcal{L}) = 0$. %

To evaluate this gradient, we also have to specify the distributional assumptions for the surrogate labels generated with $\tilde{c}$.
We assume that the conditional distribution of the surrogate labels belongs to the same EDF as the ground-truth concepts, but can be parameterized by its own distinct surrogate natural parameter $\tilde{\eta}$ and dispersion parameter $\tilde{\phi}$.
For example, if concepts are modeled as Gaussian variables, we would assume that both ground-truth and surrogate concepts are normally distributed but may exhibit a different mean and standard deviation $\tilde{\sigma}$ (where $\tilde{\phi} = \tilde{\sigma}^2$) compared to the underlying ground truth.
Substituting the corresponding gradient from \Cref{eq:gradient_mnef} into \Cref{eq:surrogate-cost-function}, we obtain:
\begin{align}
    \nabla \tilde{J}_\mathrm{task}(\tilde{\Theta};\mathcal{L}) &= \mathbb{E}_{x \sim P_{\mathrm{task}}} \left[ \nabla_\Theta \mathcal{L}(\pi_{\tilde{\Theta}}(f(x)), \tilde{c}(x)) \right]\\
    &= \mathbb{E}_{x \sim P_{\mathrm{task}}} \left[ \frac{1}{s(\tilde{\phi})} \left( \hat{c}_{\tilde{\Theta}}(x) - \tilde{c}(x) \right) \otimes f(x) \right] = 0_{K \times d}.
\end{align}
Since the dispersion scaling factor $s(\tilde{\phi})$ is strictly positive and constant with respect to the expectation, it factors out.
This implies an orthogonality condition between the surrogate residuals and the backbone activations:
\begin{equation}
    \label{eq:surrogate_orthogonality}
    \mathbb{E}_{x \sim P_{\mathrm{task}}} \left[ \left( \hat{c}_{\tilde{\Theta}}(x) - \tilde{c}(x) \right) \otimes f(x) \right] = 0_{K \times d}.
\end{equation}
Note that while we assume that $\nabla \tilde{J}_\mathrm{task}(\tilde{\Theta};\mathcal{L}) = 0_{K \times d}$, this does not imply $\tilde{J}_\mathrm{task}(\tilde{\Theta};\mathcal{L}) = 0$ (which is congruent with our empirical results in \Cref{tab:elements_faithfulness}).

\bfparagraph{5. Gradient of the True Objective.}
We define the \textit{true} objective $J^*_\mathrm{task}(\Theta;\mathcal{L})$ as the expected risk with respect to the ground-truth concept function $c^*(x)$ and the true dispersion parameter $\phi^*$:
\begin{equation}
    J^*_\mathrm{task}(\Theta;\mathcal{L}) = \mathbb{E}_{x \sim P_{\mathrm{task}}} \left[ \mathcal{L}(\pi_\Theta(f(x)), c^*(x)) \right].
\end{equation}
We evaluate the gradient of this objective at the parameters $\tilde{\Theta}$ learned via the surrogate optimization.
Substituting the gradient form from \Cref{eq:gradient_mnef}:
\begin{align}
    \nabla J^*_\mathrm{task}(\tilde{\Theta};\mathcal{L}) &= \mathbb{E}_{x \sim P_{\mathrm{task}}} \left[ \frac{1}{s(\phi^*)} \left( \hat{c}_{\tilde{\Theta}}(x) - c^*(x) \right) \otimes f(x) \right] \\
    &= \frac{1}{s(\phi^*)} \mathbb{E}_{x \sim P_{\mathrm{task}}} \left[ \left( \hat{c}_{\tilde{\Theta}}(x) - c^*(x) \right) \otimes f(x) \right] .
\end{align}
To analyze the faithfulness, we expand the residual term by adding and subtracting the surrogate label vector $\tilde{c}(x)$.
Using the linearity of the outer product and the expectation, we decompose the gradient into two components:
\begin{align}
    \nabla J^*_\mathrm{task}(\tilde{\Theta};\mathcal{L}) &= \frac{1}{s(\phi^*)} \mathbb{E}_{x \sim P_{\mathrm{task}}} \left[ \left( \hat{c}_{\tilde{\Theta}}(x) - \tilde{c}(x) + \tilde{c}(x) - c^*(x) \right) \otimes f(x) \right] \\
    &= \frac{1}{s(\phi^*)} \Bigg( \underbrace{\mathbb{E}_{x \sim P_{\mathrm{task}}} \left[ (\hat{c}_{\tilde{\Theta}}(x) - \tilde{c}(x)) \otimes f(x) \right]}_{0 \text{ via surrogate optimality (\Cref{eq:surrogate_orthogonality})}} \nonumber \\
    &\quad\quad\quad\quad + \mathbb{E}_{x \sim P_{\mathrm{task}}} \left[ (\tilde{c}(x) - c^*(x)) \otimes f(x) \right] \Bigg).
\end{align}
The first term vanishes because $\tilde{\Theta}$ is a stationary point of the surrogate objective.
Consequently, the gradient of the true objective depends only on the second term:
\begin{equation}
    \label{eq:final_alignment}
    \nabla J^*_\mathrm{task}(\tilde{\Theta};\mathcal{L}) = \frac{1}{s(\phi^*)} \mathbb{E}_{x \sim P_{\mathrm{task}}} \left[ \underbrace{\left( \tilde{c}(x) - c^*(x) \right)}_{\text{Label Discrepancy}} \otimes f(x) \right].
\end{equation}
Now, $J^*_\mathrm{task}(\tilde{\Theta};\mathcal{L})$ depends only on the discrepancy between the true concept labels from $c^*$ and the surrogate concept labels generated with $\tilde{c}$, not on the specific concept projection $\pi_{\tilde{\Theta}}$.

This result provides a mathematical condition for the surrogate faithfulness.
The surrogate parameters $\tilde{\Theta}$ are optimal for the true objective if and only if this gradient (\Cref{eq:final_alignment}) is the zero matrix.
This condition can be satisfied in two distinct ways.
First, and most intuitively, if the surrogate labels are perfectly accurate ($\tilde{c}(x) = c^*(x)$ for all $x$), the discrepancy term vanishes, and the gradient is zero.
Second, and more subtly, the model remains faithful even with imperfect labels if the label discrepancy is \emph{orthogonal} to the activation space in expectation.
For the gradient matrix to be zero, each of its elements must be zero.
Let $\delta(x) = \tilde{c}(x) - c^*(x)$ be the label discrepancy vector.
The condition for the $(k, j)$-th element is:
\begin{equation}
    \label{eq:orthogonality_elementwise}
    \mathbb{E}_{x \sim P_{\mathrm{task}}} \left[ \delta_k(x) \cdot f_j(x) \right] = 0, \quad \forall k \in \{1,\dots,K\}, j \in \{1,\dots,d\}.
\end{equation}
To build intuition for this condition, we can consider its empirical approximation over the $N$ samples of the training set, as shown in Figure~\ref{fig:orthogonality_comparison}.
Here, it requires the dot product of every two vectors in the $N$-dimensional sample space to be zero: the vector of label discrepancies for concept $k$, $\mathbf{\Delta}_k = [\delta_k(x_1), \dots, \delta_k(x_N)]^\top$, and the vector of feature values for dimension $j$, $\boldsymbol{\alpha}_j = [f_j(x_1), \dots, f_j(x_N)]^\top$.
Therefore, faithfulness does not strictly require accurate surrogate labels; it only requires that the errors in these labels are not systematically correlated with the features the backbone model has learned.

\end{appendixblock}

\section{Additional Experimental Details}
\subsection{Experimental Setup: Random Concept Projections}
\label{apx:random-concepts}

\begin{appendixblock}

As motivated in \Cref{sec:random-exp}, we empirically investigate whether downstream predictive accuracy is a sufficient metric for evaluating the quality of a Post-Hoc Concept Bottleneck Model (PCBM)~\cite{yuksekgonul2023posthoc}.
Our theoretical analysis in \Cref{sec:method-downstream} suggests that even using semantically uninformative random projections for the concept projection $\pi_{\theta}$ can preserve enough information to enable high task performance.
To empirically validate our theoretical results, we examine the performance of a PCBM equipped with a randomly generated concept projection, following a setup similar to that of \cite{midavaine2024reproducibility}.
If a meaningless bottleneck can yield high predictive accuracy, accuracy alone is an unreliable proxy for concept faithfulness.

\paragraph{Concept Projection.}
Generally, PCBMs implement the concept projection $\pi_{\theta}: \mathcal{A} \to \mathcal{C}$ mapping from $d$-dimensional backbone activations $a \in \mathcal{A} \subseteq \mathbb{R}^d$ to a $K$-dimensional concept space $\mathcal{C} \subseteq \mathbb{R}^K$ as a linear layer with a potentially non-linear activation function (\Cref{apx:surrogate-faithfulness-derivation}).
In this experiment, we follow the standard PCBM procedure~\cite{yuksekgonul2023posthoc} and implement $\pi_{\theta}$ as a linear transformation of the backbone activations (i.e., selecting identity transformation as activation function).
We further follow \cite{yuksekgonul2023posthoc} and set the bias vector $\mathbf{b} \in \mathbb{R}^K$ to $\mathbf{0}$.
Consequently, $\pi_{\theta}$ is parameterized only by a weight matrix $\Theta \in \mathbb{R}^{K \times d}$ and implements the following mapping for backbone activations $a \in \mathcal{A}$:
\begin{equation}
    \pi_{\Theta}(a) = \Theta a.
\end{equation}

\paragraph{Construction of Random Concept Projections.}
To construct a semantically meaningless concept projection, we generate a random weight matrix $\Theta_{\mathrm{rand}} \in \mathbb{R}^{K \times d}$ in a two-step process following our theoretical results presented in \Cref{sec:method-downstream}.
First, we sample an initial random matrix $\Theta'_{\mathrm{rand}}$ where each element follows a standard normal distribution:
\begin{equation}
    \Theta'_{\mathrm{rand}} \sim \mathcal{N}(0, 1)^{K \times d}.
\end{equation}
Afterward, we perform $L_2$-normalization of each row vector of $\Theta'_{\mathrm{rand}}$ to obtain the final weight matrix $\Theta_{\mathrm{rand}}$ (similarly to CAVs~\cite{kim2018interpretability}).
Since the transformation matrix $\Theta_{\mathrm{rand}}$ consists of unit-length basis vectors with random direction, the resulting random concept projection $\pi_{\Theta_{\mathrm{rand}}}(a) = \Theta_{\mathrm{rand}} a$ maps the backbone activations $a \in \mathcal{A}$ into a semantically meaningless concept space.

\paragraph{Model Architecture and Training.}
While in the standard PCBM framework~\cite{yuksekgonul2023posthoc} only the backbone model $f: \mathcal{X} \to \mathcal{A}$ is frozen, we now also freeze the concept projection, implementing it as the random concept projection $\pi_{\Theta_{\mathrm{rand}}}$ defined above.
Thus, the downstream classifier $h: \mathcal{C} \to \mathcal{Y}$ is the only trainable component, optimized on the outputs of $\pi_{\Theta_{\mathrm{rand}}}$.
Following \cite{oikarinen2023label}, we implement $h$ as a single linear layer trained with a cross-entropy loss and elastic-net regularization \cite{zou2005elasticnet}.
This regularization encourages sparsity in the concept-to-class weights, which is commonly desired for the interpretability of the classifier head.
The backbone architectures and bottleneck insertion points are adapted for each dataset:
\begin{itemize}
    \item \textbf{CUB~\cite{welinder2010cub}:} We follow the standard train-test split~\cite{welinder2010cub} and use a ResNet-18~\cite{he2016resnet} pretrained on ImageNet~\cite{deng2009imagenet} for the backbone $f$. The concept bottleneck is inserted after the final global average pooling layer (\texttt{features.final\_pool}). The classifier head $h$ is trained with the Adam optimizer \cite{kingma2014adam} at a learning rate of $2 \times 10^{-4}$.
    \item \textbf{Elements~\cite{nicolson2025explaining}:} Following \cite{nicolson2025explaining}, we select a simple $5$-layer convolutional neural network (ConvNet) as backbone, which uses $64$ channels per block. The bottleneck is inserted after the third convolutional layer (\texttt{layer3}). The classifier $h$ is trained with Adam~\cite{kingma2014adam} at a learning rate of $1 \times 10^{-5}$.
    \item \textbf{CIFAR-10 \& CIFAR-100~\cite{krizhevsky2009cifar}:} We follow the default train-test split~\cite{krizhevsky2009cifar}. The backbone is a ResNet-50~\cite{he2016resnet} pretrained via CLIP~\cite{radford2021learning}, where we insert the concept bottleneck after the final residual block (\texttt{layer4}).
    The classifier $h$ is trained with Adam~\cite{kingma2014adam} at a learning rate of $7 \times 10^{-4}$.
\end{itemize}

\paragraph{Reconstruction of Backbone Activations.}
To directly quantify the information preserved by the random concept projection $\pi_{\Theta_{\mathrm{rand}}}$, we train a decoder to reconstruct the original backbone activations.
We define a linear decoder $\mathcal{D}: \mathcal{C} \to \mathcal{A}$, parameterized by a weight matrix $\Theta_{\mathcal{D}} \in \mathbb{R}^{d \times K}$, which is trained to minimize the Mean Squared Error (MSE) between the original activations $a \in \mathcal{A}$ and the reconstructed activations $\hat{a} = \mathcal{D}(\pi_{\Theta_{\mathrm{rand}}}(a))$.
A low reconstruction MSE indicates that the concept space, despite being generated randomly, retains sufficient information to recover the original activations.
This validates our theoretical argument that a sufficiently expressive downstream classifier $h$ can learn to approximate this inverse mapping, thereby bypassing any semantic interpretation of the concepts to achieve high task accuracy.

\end{appendixblock}

\subsection{Additional Results: Random Concept Projections}
\label{apx:random-concept-add-results}

\begin{appendixblock}

In this section, we provide extended results for the experiments on random concept projections introduced in \Cref{sec:random-exp}, with setup details available in \Cref{apx:random-concepts}.
\Cref{fig:three_stacked_images} visualizes the task classification accuracy of PCBMs \cite{yuksekgonul2023posthoc} with concept projections trained on the CUB \cite{welinder2010cub}, CIFAR-10 \cite{krizhevsky2009cifar}, and Elements \cite{nicolson2025explaining} datasets and compares them with PCBMs utilizing a semantically meaningless concept projection $\pi_{\Theta_{\mathrm{rand}}}$ for the respective dataset (\Cref{apx:random-concepts}).
For the concept projection $\pi_{\Theta_{\mathrm{rand}}}$, we scale the bottleneck dimension $K$ of the weight matrix $\Theta_{\mathrm{rand}} \in \mathbb{R}^{K \times d}$ to align with the number of concepts present in the trained PCBMs.
The resulting baseline curve extends slightly past the maximum number of concepts used by any PCBM for the task.
For the trained concept projections $\pi_{\theta}$, we use auxiliary data (Broden~\cite{bau2017network}) and VLM-based surrogate methods (LFCBM~\cite{oikarinen2023label} and VLG-CBM~\cite{srivastava2024vlg}, see \Cref{apx:vlm-surrogate-functions}).
The concepts for the VLM-based surrogate methods are sourced from ground-truth concept labels, ConceptNet 5.5~\cite{speer2017conceptnet}, and GPT-3.5~\cite{brown2020language}.

Consistent with the findings presented in the main text (\Cref{sec:random-exp}), we observe that as the dimensionality $K$ of the random concept projection $\pi_{\Theta_{\mathrm{rand}}}$ increases, its downstream test accuracy asymptotically approaches that of an unconstrained base classifier without a bottleneck layer.
Additionally, the PCBMs with learned concept projections only marginally outperform the random concept projections of equivalent dimensionality $K$, indicating that task accuracy is an inadequate metric to measure the semantic meaningfulness of a PCBM.
To investigate why random concept projections can achieve this classification performance, we additionally report the mean squared error (MSE) for reconstructing the $d$-dimensional backbone activations from the $K$-dimensional outputs of the random projection (\Cref{apx:random-concepts}).
We observe that as $K$ increases, the reconstruction MSE monotonically decreases, suggesting that higher-dimensional random projections retain sufficient information about the backbone activations for the downstream classifier $h$ to solve the task directly (c.f., \Cref{sec:method-downstream}).
Consequently, these results highlight that downstream task accuracy is an insufficient metric to assess whether a PCBM relies on semantically meaningful concepts.

\begin{figure}[htbp]
    \centering

    \begin{subfigure}{\textwidth}
        \centering
        \includegraphics[width=\linewidth]{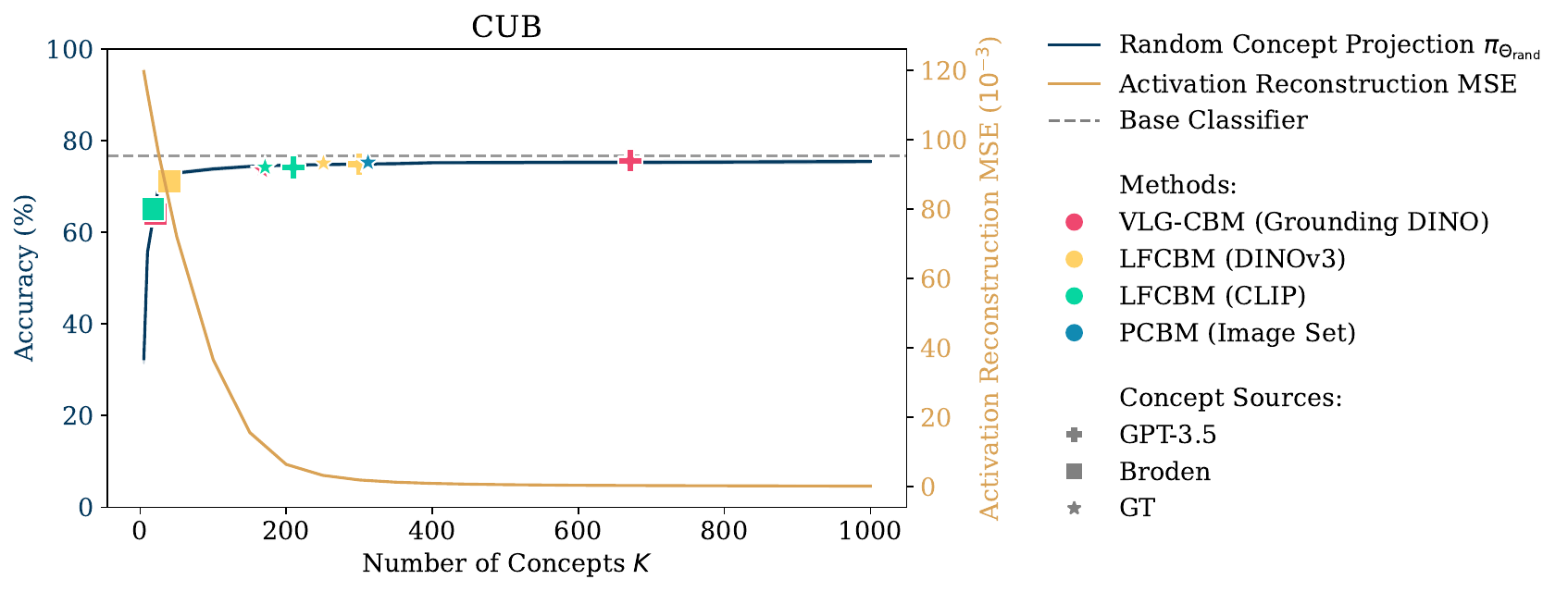}
        \caption{CUB}
        \label{fig:top}
    \end{subfigure}

    \vspace{0.3cm}

    \begin{subfigure}{\textwidth}
        \centering
        \includegraphics[width=\linewidth]{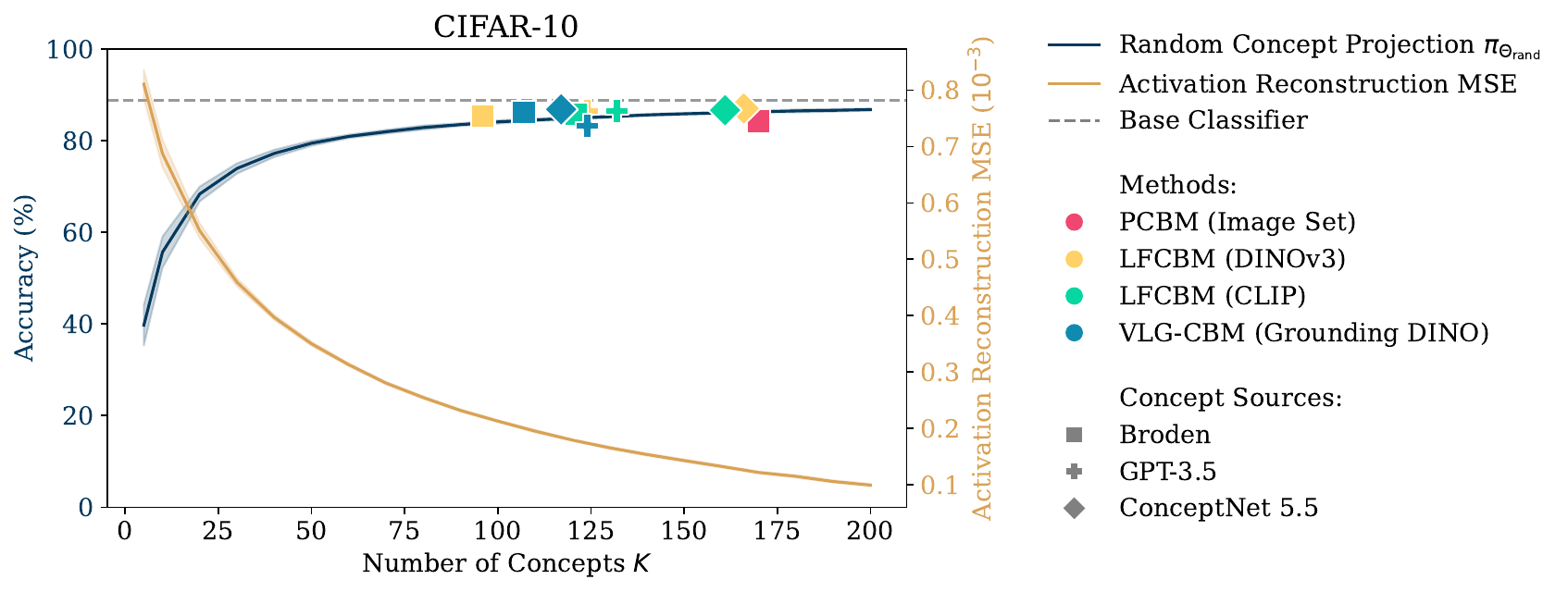}
        \caption{CIFAR10}
        \label{fig:middle}
    \end{subfigure}

    \vspace{0.3cm}

    \begin{subfigure}{\textwidth}
        \centering
        \includegraphics[width=\linewidth]{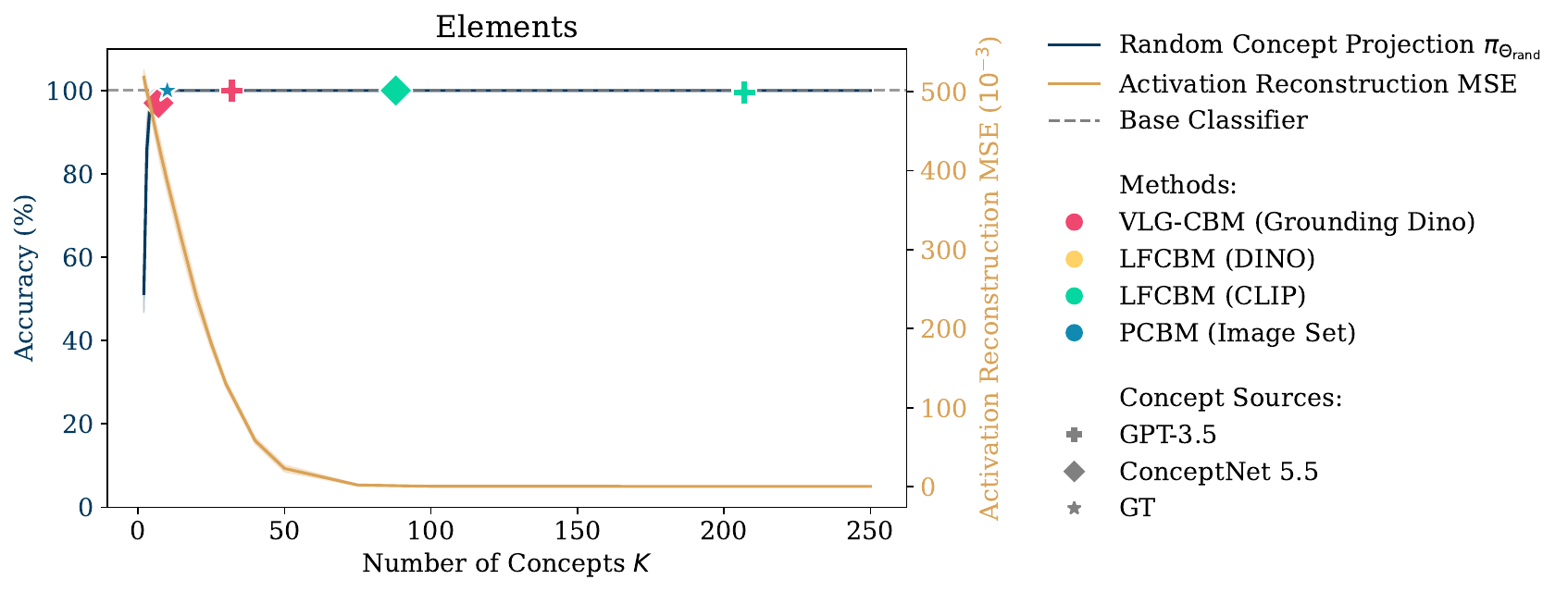}
        \caption{Elements}
        \label{fig:bottom}
    \end{subfigure}

    \caption{Extended results for the random concept projection experiments in \Cref{fig:random_directions_accuracy} on (a) CUB \cite{welinder2010cub}, (b) CIFAR-10 \cite{krizhevsky2009cifar}, and (c) Elements \cite{nicolson2025explaining}.
    As the dimensionality $K$ of $\pi_{\Theta_{\mathrm{rand}}}$ increases, the downstream task accuracy (blue line) approaches that of an unconstrained base classifier without a bottleneck layer (dashed line).
    This is consistent with a corresponding drop in activation reconstruction MSE (orange line), indicating that higher-dimensional random projections retain sufficient information to reconstruct the backbone activations.
    The PCBMs with random concept projections also achieve a performance similar to that of standard PCBMs of similar dimensionality, where $\pi_{\theta}$ is trained on an auxiliary concept set or VLM-based surrogate labels (scatter points).
    }
    \label{fig:three_stacked_images}
\end{figure}

\end{appendixblock}

\subsection{Measuring Alignment of Concept Activation Vectors}
\label{apx:alignment-cav}

\begin{appendixblock}
To quantify the \emph{geometric faithfulness} of the concept transformation $\pi_{\theta}$, we compare the learned concept directions with the ground truth.
For CAV-based concept transformations~\cite{kim2018interpretability}, this is straightforward.
Here, we measure the geometric faithfulness by extracting the concept activation vectors $v_k$ (for concept $k$) and computing their alignment with the ground-truth directions $v_k^*$ (obtained by training on the full training dataset $X_{\mathrm{task}}$ with known ground-truth concept labels) via cosine similarity:
\begin{equation}
    \mathrm{sim}_\sphericalangle(v_k, v_k^*) = \frac{\langle v_k, v_k^* \rangle}{\| v_k \| \| v_k^* \|}. %
\end{equation}
Since both LFCBMs~\cite{oikarinen2023label} and VLG-CBMs~\cite{srivastava2024vlg} implement $\pi_{\theta}$ as a \emph{Generalized Linear Model}~\cite{Fahrmeir.2001,McCullagh.1989} (\Cref{apx:surrogate-faithfulness-derivation}), they follow the form:
\begin{equation}
    \pi_\Theta(f(x)) = \psi^{-1}(\Theta f(x)),
\end{equation}
where $\psi$ denotes the link function of the GLM and $\Theta \in \mathbb{R}^{K \times d}$ is a learned weight matrix.
Thus, we consider the normalized row vectors of $\Theta$ as our concept activation vectors $v_k$ and calculate $\mathrm{sim}_\sphericalangle$ analogously to the CAV-based PCBMs. 

\end{appendixblock}

\subsection{Additional Results: Alternative Faithfulness Metrics}
\label{apx:alternative-faithfulness-metrics-results}

\begin{appendixblock}

\begin{table}[t]
    \centering
    \caption{Information leakage following \cite{makonnen2025measuring,schoen2025measuring}. 
    We highlight the generalization error on $P_\mathrm{task}$ (\Cref{eq:faithful}) in \textbf{bold}.
    Note that NICC~\cite{schoen2025measuring} for the optimal $h^*$ is zero by definition (marked by $^\dagger$).
    }
    \label{tab:leakage}
    \setlength{\tabcolsep}{6pt}
    \begin{tabular}{llcccc}
    \toprule
    Method & $P_{\mathrm{aux}}$ &\textbf{ $\boldsymbol{P}_\mathrm{\textbf{task}}$ Error $\boldsymbol{[\downarrow]}$} & \multicolumn{2}{c}{NICC $[\downarrow]$~\cite{schoen2025measuring}} & $I(y;\hat{c},c)$ $[\downarrow]$~\cite{makonnen2025measuring}\\
    \cmidrule(lr){4-5}
    & & & $h$&$h^*$ & \\
    \midrule
    \multirow{4}{*}{PCBM~\cite{yuksekgonul2023posthoc}}
     & $P_{\mathrm{task}}$ & 0.0000{\scriptsize $\pm$0.0000} & 0.5563{\scriptsize $\pm$0.0038} & 0.0$^\dagger$& 0.0040{\scriptsize $\pm$0.0002}\\
     & $P_{\mathrm{near}}^{\forall}$ & 0.5000{\scriptsize $\pm$0.0076}  & 0.4335{\scriptsize $\pm$0.0065} & 0.0$^\dagger$& 0.0040{\scriptsize $\pm$0.0002}\\
     & $P_{\mathrm{near}}^{\exists}$ & 0.3225{\scriptsize $\pm$0.0030}  & 0.4909{\scriptsize $\pm$0.0144} & 0.0$^\dagger$& 0.0047{\scriptsize $\pm$0.0003}\\
     & $P_{\mathrm{OOD}}$ & 0.4852{\scriptsize $\pm$0.0019}  & 0.4105{\scriptsize $\pm$0.0130} & 0.0$^\dagger$& 0.0048{\scriptsize $\pm$0.0002} \\

    \bottomrule
    \end{tabular}

    \vspace{-0.4cm}
\end{table}

Unlike information leakage approaches (e.g., NICC~\cite{schoen2025measuring} and \cite{makonnen2025measuring}) that assess the interplay between $\pi_\theta$ and the downstream classifier $h$, our definition of faithfulness strictly isolates $\pi_\theta$. 
These perspectives are complementary: unfaithfulness can occur without leakage, and conversely, for a faithful $\pi_\theta$, an optimized $h$ may still exploit leaked information.
This is demonstrated in \Cref{tab:leakage}, which evaluates the information leakage metrics presented by~\cite{schoen2025measuring,makonnen2025measuring} and reveals differences with respect to our approach.
Row 1 shows $\pi_\theta$ trained on $P_\mathrm{task}$ minimizing the associated error, yet $h$ exploits leaked information (NICC $> 0$). 
In contrast, for the specified optimal $h^*$, NICC reports zero leakage even if $P_\mathrm{aux} \neq P_\mathrm{task}$, leading to unfaithful $\pi_\theta$.

\end{appendixblock}

\subsection{Additional Results: Covariate Shift}
\label{apx:covariate-shift-results}

\begin{algorithm}[htbp]
\caption{Estimation of $\mathcal{H}\Delta\mathcal{H}$-Divergence and Generalization Error}
\label{alg:cov_shift_experiment}
\begin{algorithmic}[1]
\Require Set of $M$ distributions $\mathcal{D} = \{P_1, \dots, P_M\}$, set of concepts $\mathcal{K} = \{1, \dots, K\}$, backbone network $f: \mathcal{X} \to \mathcal{A}$, and oracle concept labeling function $c^*: \mathcal{X} \to [0,1]^K$.

\State Initialize results matrices $\mathbf{H}, \mathbf{E} \in \mathbb{R}^{M \times M}$

\For{$s = 1$ to $M$ and $t = 1$ to $M$}
    \State Set $P_{\mathrm{source}} \gets P_s$, $P_{\mathrm{target}} \gets P_t$
    \State Initialize cumulative divergence $D \gets 0$, cumulative error $E \gets 0$
    
    \For{$k = 1$ to $K$}
        \Statex
        \markcomment{2}{\textbf{1. Data Sampling and Preparation}}
        \State $\mathbb{X}^+ \sim P_{\mathrm{source}}(\cdot \mid c_k = 1)$
        \State $\mathbb{X}^- \sim P_{\mathrm{target}}(\cdot \mid c_k = 0)$ 
        \State $\mathbb{X}_{\mathrm{task}} \sim P_{\mathrm{target}}$ %
        
        \State $\mathbb{X}_{\mathrm{aux}} \gets \{(x, 1) \mid x \in \mathbb{X}^+\} \cup \{(x, 0) \mid x \in \mathbb{X}^-\}$
        \State $\mathbb{X}_{\mathrm{aux}}^{\mathrm{train}}, \mathbb{X}_{\mathrm{aux}}^{\mathrm{test}} \gets \mathrm{Split}(\mathbb{X}_{\mathrm{aux}})$
        \State $\mathbb{X}_{\mathrm{task}}^{\mathrm{train}}, \mathbb{X}_{\mathrm{task}}^{\mathrm{test}} \gets \mathrm{Split}(\mathbb{X}_{\mathrm{task}})$

        \Statex
        \markcomment{2}{\textbf{2. Concept Classifier Optimization}}
        \State $\mathbb{A}_{\mathrm{aux}}^{\mathrm{train}} \gets \{ (f(x), y) \mid (x, y) \in \mathbb{X}_{\mathrm{aux}}^{\mathrm{train}} \}$
        \State $\pi_\theta^{(k)} \gets \mathrm{TrainCAV}(\mathbb{A}_{\mathrm{aux}}^{\mathrm{train}})$

        \Statex
       \markcomment{2}{\textbf{3. Domain Discriminator Optimization and Evaluation}}
        \State $\mathbb{A}_{\mathrm{disc}}^{\mathrm{train}} \gets \{ (f(x), 0) \mid (x, y) \in \mathbb{X}_{\mathrm{aux}}^{\mathrm{train}} \} \cup \{ (f(x), 1) \mid x \in \mathbb{X}_{\mathrm{task}}^{\mathrm{train}} \}$
        \State $h_{\mathrm{disc}} \gets \mathrm{TrainSGDClassifier}(\mathbb{A}_{\mathrm{disc}}^{\mathrm{train}})$
        
        \State $\mathbb{A}_{\mathrm{disc}}^{\mathrm{test}} \gets \{ (f(x), 0) \mid (x, y) \in \mathbb{X}_{\mathrm{aux}}^{\mathrm{test}} \} \cup \{ (f(x), 1) \mid x \in \mathbb{X}_{\mathrm{task}}^{\mathrm{test}} \}$
        \State $\epsilon_{\mathrm{disc}}^{(k)} \gets \mathrm{Error}(h_{\mathrm{disc}}, \mathbb{A}_{\mathrm{disc}}^{\mathrm{test}})$
        \State $\hat{d}_{\mathcal{H}\Delta\mathcal{H}}^{(k)} \gets 2(1 - 2\epsilon_{\mathrm{disc}}^{(k)})$ \Comment{Approximation following \cite{BenDavid.2010}}
        \State $D \gets D + \hat{d}_{\mathcal{H}\Delta\mathcal{H}}^{(k)}$

        \Statex
        \markcomment{2}{\textbf{4. Generalization Error Calculation}}
        \State $\mathbb{A}_{\mathrm{task}}^{\mathrm{test}} \gets \{ (f(x), c_k^*(x)) \mid x \in \mathbb{X}_{\mathrm{task}}^{\mathrm{test}} \}$ \Comment{Add oracle labels}
        \State $\mathbb{A}_{\mathrm{aux}}^{\mathrm{test}} \gets \{ (f(x), y) \mid (x, y) \in \mathbb{X}_{\mathrm{aux}}^{\mathrm{test}} \}$
        \State $\epsilon_{\mathrm{gen}}^{(k)} \gets \left| \mathrm{Error}(\pi_\theta^{(k)}, \mathbb{A}_{\mathrm{task}}^{\mathrm{test}}) - \mathrm{Error}(\pi_\theta^{(k)}, \mathbb{A}_{\mathrm{aux}}^{\mathrm{test}}) \right|$
        \State $E \gets E + \epsilon_{\mathrm{gen}}^{(k)}$
    \EndFor
    \State Record averages: $\mathbf{H}_{s,t} \gets D/K$, \quad $\mathbf{E}_{s,t} \gets E/K$
\EndFor
\Statex
\State \Return $\mathbf{H}$ (Pairwise Divergences), $\mathbf{E}$ (Pairwise Generalization Errors)
\end{algorithmic}
\end{algorithm}

We extend the covariate shift experiments conducted on the synthetic Elements dataset~\cite{nicolson2025explaining} (\Cref{sec:covariate-shift-exp}) to the real-world CUB dataset~\cite{welinder2010cub}.
Specifically, we compare the impact of training the concept projection $\pi_{\theta}$ on auxiliary concept sets from Broden~\cite{bau2017network} to training it on in-domain CUB concept sets.
After mapping the aligned concepts between the two datasets (e.g., matching ``Bill'' in CUB to ``Beak'' in Broden, and ``Upperparts'' to ``Body''), we can leverage CUB's ground-truth annotations to compute the true generalization error of the $\pi_{\theta}$ and use it to validate our proposed label-free faithfulness metric: the estimated $\mathcal{H}\Delta\mathcal{H}$-divergence~\cite{BenDavid.2010}, which bounds the generalization error (\Cref{sec:method-concept-projection}).

To reflect realistic post-hoc interpretability scenarios in which concept annotations for the target domain are missing, we draw only negative concept examples from the target domain and restrict the sampling of positive examples to the auxiliary domain.
As detailed in \Cref{alg:cov_shift_experiment}, we construct an auxiliary training set $\mathbb{X}_{\mathrm{aux}}$ for each concept $k$ by combining positive samples from the source distribution with negative samples from the target distribution.
The concept projection $\pi_\theta$ is then trained per-concept on $\mathbb{X}_{\mathrm{aux}}$.
To prevent data leakage caused by the limited sample sizes, we strictly enforce that no input sample appears in both the training and test sets.

Next, we calculate our faithfulness metric for distribution shift by training a linear domain discriminator on the backbone features to separate $\mathbb{X}_{\mathrm{aux}}$ from the target distribution (\Cref{sec:method-concept-projection}).
Following \cite{BenDavid.2010}, we optimize it using a Huber loss~\cite{Zhang.2004}, and approximate the $\mathcal{H}\Delta\mathcal{H}$-divergence as $\hat{d}_{\mathcal{H}\Delta\mathcal{H}} \approx 2(1 - 2\epsilon)$ based on the discriminator's classification error $\epsilon$.
We execute this procedure for all pairwise combinations of CUB and Broden acting as source and target domains. 

\begin{figure}[htbp]
    \centering
    \begin{subfigure}[t]{0.48\textwidth}
        \centering
        \includegraphics[width=\linewidth]{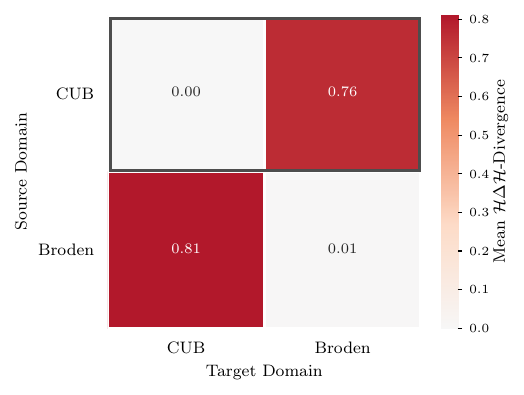}
        \caption{Average Pairwise $\mathcal{H}\Delta\mathcal{H}$-Divergence Matrix}
        \label{fig:pairwise-h-divergences-cub}
    \end{subfigure}
    \hfill
    \begin{subfigure}[t]{0.48\textwidth}
        \centering
        \includegraphics[width=\linewidth]{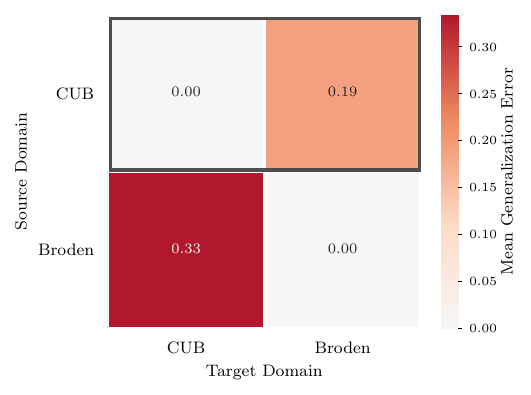}
        \caption{Pairwise Generalization Errors}
        \label{fig:pairwise-generalization-errors-cub}
    \end{subfigure}
    \caption{Visualization of the estimated $\mathcal{H}\Delta\mathcal{H}$-divergence~\cite{BenDavid.2010} as a proxy for the upper bound of the task generalization error for the CUB dataset~\cite{welinder2010cub}, when comparing training $\pi_{\theta}$ based on the ground-truth concept labels to training it on the Broden dataset~\cite{bau2017network}.
    As in the main text, stronger distribution shifts between the concept-probing datasets and the downstream domain, as indicated by our divergence metric, indicate greater unfaithfulness, as observed from the true generalization error.
    By estimating the $\mathcal{H}\Delta\mathcal{H}$-divergence, we can identify this degradation in faithfulness without requiring target domain concept annotations.
    }
    \label{fig:H-divergence-cub}
\end{figure}

The resulting divergence estimates and corresponding true generalization errors are presented in \Cref{fig:H-divergence-cub}.
Consistent with our findings on the Elements dataset, we observe a strong positive Pearson correlation~\cite{pearson1896correlation} ($r > 0.85$) between our divergence metric and the actual generalization error.
This confirms that relying on visually distinct auxiliary datasets, such as Broden, introduces a measurable covariate shift that explicitly degrades the faithfulness of the learned concept representations.
Furthermore, it demonstrates that approximating the generalization error by the empirical $\mathcal{H}\Delta\mathcal{H}$-divergence metric successfully identifies this degradation without requiring target domain concept annotations.

\subsection{Additional Results: Systematic Surrogate Label Errors}
\label{apx:systematic-surrogate-errors-results}

\begin{appendixblock}

In this section, we expand the analysis of systematic surrogate concept label errors by evaluating additional Vision-Language Models (VLMs; \Cref{apx:vlm-surrogate-functions}).
Following \Cref{sec:method-concept-projection}, we compute the Pearson correlation~\cite{pearson1896correlation} between the backbone activations and the surrogate errors produced by CLIP~\cite{radford2021learning}, DINOv3~\cite{simeoni2025dinov3}, and Grounding DINO~\cite{liu2024grounding}.
We compare these results against a synthetic baseline of unsystematic label noise, implemented by randomly flipping the binary concept labels with a 40\% probability.

\Cref{fig:four_graphs_elements_correlation} presents the Pearson correlation coefficients and average absolute surrogate errors for the Elements dataset~\cite{nicolson2025explaining}.
To highlight the systematic nature of the VLM-induced errors, \Cref{fig:four_graphs_elements_pvalue} masks the correlation matrix, displaying only statistically significant entries ($p < 0.05$).
We repeat the same experiment for the real-world CUB dataset~\cite{welinder2010cub}, presenting the unmasked correlations in \Cref{fig:four_graphs_cub_correlation} and the masked significant correlations in \Cref{fig:four_graphs_cub_pvalue}.
To compute the surrogate label error for CUB, we use the human-annotated ground-truth attributes and analyze a subset of the first $50$ concepts.
Across all evaluated VLMs and datasets, the surrogate label errors exhibit statistically significant correlations with the backbone activations.
On the other hand, the 40\% random-noise baseline demonstrates no such systematic structure.
These results further validate that the current VLM-based generation of surrogate labels can violate the error orthogonality requirement, introducing systematic errors into the concept bottleneck.

\begin{figure}[htbp]
     \centering
     \begin{minipage}{0.86\textwidth}
     \begin{subfigure}[b]{0.48\textwidth}
         \centering
         \includegraphics[width=\textwidth]{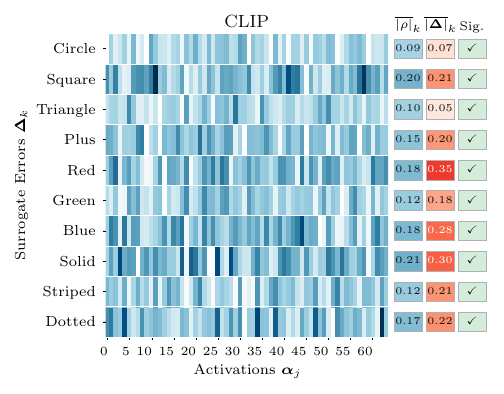}
         \caption{CLIP}
         \label{fig:image1}
     \end{subfigure}
     \hfill
     \begin{subfigure}[b]{0.48\textwidth}
         \centering
         \includegraphics[width=\textwidth]{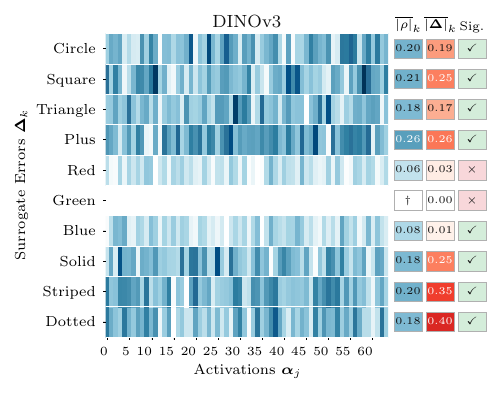}
         \caption{DINOv3}
         \label{fig:image2}
     \end{subfigure}

     \begin{subfigure}[b]{0.48\textwidth}
         \centering
         \includegraphics[width=\textwidth]{figures/single-simple-elements_vlg_cbm_pearson_correlation_heatmap.pdf}
         \caption{Grounding DINO}
         \label{fig:image3}
     \end{subfigure}
     \hfill
     \begin{subfigure}[b]{0.48\textwidth}
         \centering
         \includegraphics[width=\textwidth]{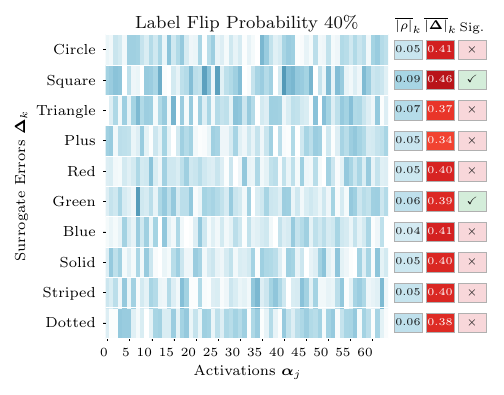}
         \caption{Random Label Noise}
         \label{fig:image4}
     \end{subfigure}
     \end{minipage}
     \hfill
     \begin{minipage}{0.10\textwidth}
         \centering
         \includegraphics[width=\textwidth, trim={215 0 0 0}, clip]{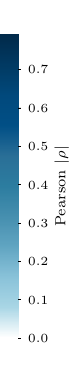}
     \end{minipage}
     \caption{Systematic vs. unsystematic surrogate errors for various VLMs (\Cref{apx:vlm-surrogate-functions}), extending the results from \Cref{fig:pearson_correlation_heatmap}.
     We visualize the Pearson correlation~\cite{pearson1896correlation} between backbone activations and surrogate label errors on the Elements dataset~\cite{nicolson2025explaining}.
     Concept labels generated by (a) CLIP~\cite{radford2021learning}, (b) DINOv3~\cite{simeoni2025dinov3}, and (c) Grounding DINO~\cite{liu2024grounding} all exhibit systematic, correlated errors (visible as dense vertical bands).
     In contrast, randomly flipping 40\% of the concept labels (d) yields an unsystematic error profile, satisfying the error orthogonality condition (\Cref{sec:method-concept-projection}).
     If a concept is predicted without error ($\overline{|\mathbf{\Delta}|}_k = 0$), the correlation coefficient is undefined ($\dagger$).
     }
     \label{fig:four_graphs_elements_correlation}
\end{figure}

\begin{figure}[htbp]
    \centering
     \begin{minipage}{0.86\textwidth}
     \begin{subfigure}[b]{0.48\linewidth}
         \centering
         \includegraphics[width=\textwidth, trim={0 0 0 16}, clip]{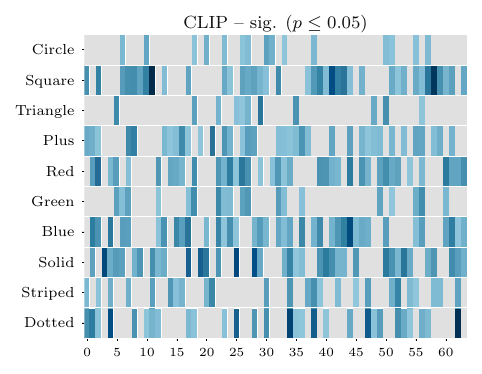}
         \caption{CLIP}
         \label{fig:sys_noise_clip_elements_masked}
     \end{subfigure}
     \hfill
     \begin{subfigure}[b]{0.48\textwidth}
         \centering
         \includegraphics[width=\textwidth, trim={0 0 0 16}, clip]{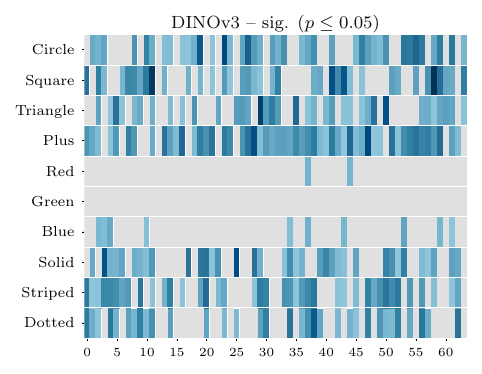}
         \caption{DINOv3}
         \label{fig:sys_noise_dino_elements_masked}
     \end{subfigure}

     \begin{subfigure}[b]{0.48\textwidth}
         \centering
         \includegraphics[width=\textwidth, trim={0 0 0 16}, clip]{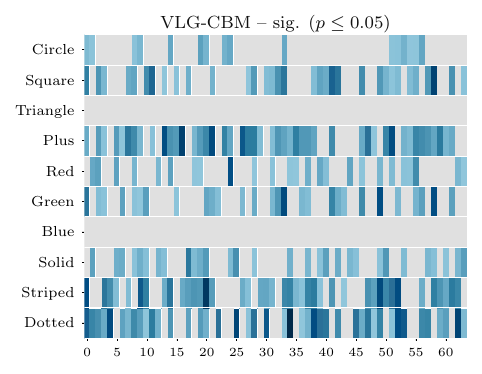}
         \caption{Grounding DINO}
         \label{fig:sys_noise_vlg_elements_masked}
     \end{subfigure}
     \hfill
     \begin{subfigure}[b]{0.48\textwidth}
         \centering
         \includegraphics[width=\textwidth, trim={0 0 0 16}, clip]{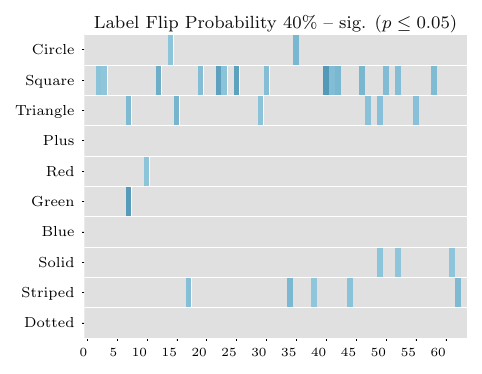}
         \caption{Random Label Noise}
         \label{fig:random_noise_elements_masked}
     \end{subfigure}
     \end{minipage}
     \hfill
     \begin{minipage}{0.10\textwidth}
         \centering
         \includegraphics[width=\textwidth, trim={215 0 0 0}, clip]{figures/pearson_correlation_coefficient_legend.pdf}
     \end{minipage}
     \caption{
     Statistically significant Pearson correlation coefficients~\cite{pearson1896correlation} from \Cref{fig:four_graphs_elements_correlation} (surrogate errors on the Elements dataset~\cite{nicolson2025explaining}).
     We display only correlation coefficients that remain significant after a Holm-Bonferroni correction~\cite{holm1979correction} ($p < 0.05$).
     Surrogate labels generated by (a) CLIP~\cite{radford2021learning}, (b) DINOv3~\cite{simeoni2025dinov3}, and (c) Grounding DINO~\cite{liu2024grounding} exhibit dense blocks of statistically significant correlations, confirming that their labeling errors are systematic.
     Conversely, the random noise baseline (d) shows no significant correlations beyond the expected false positive rate.
    }
    \label{fig:four_graphs_elements_pvalue}
\end{figure}
    
\begin{figure}[htbp]
     \centering
     \begin{minipage}{0.86\textwidth}
     \begin{subfigure}[t]{0.48\textwidth}
         \centering
         \includegraphics[width=\textwidth,height=0.36\textheight,keepaspectratio]{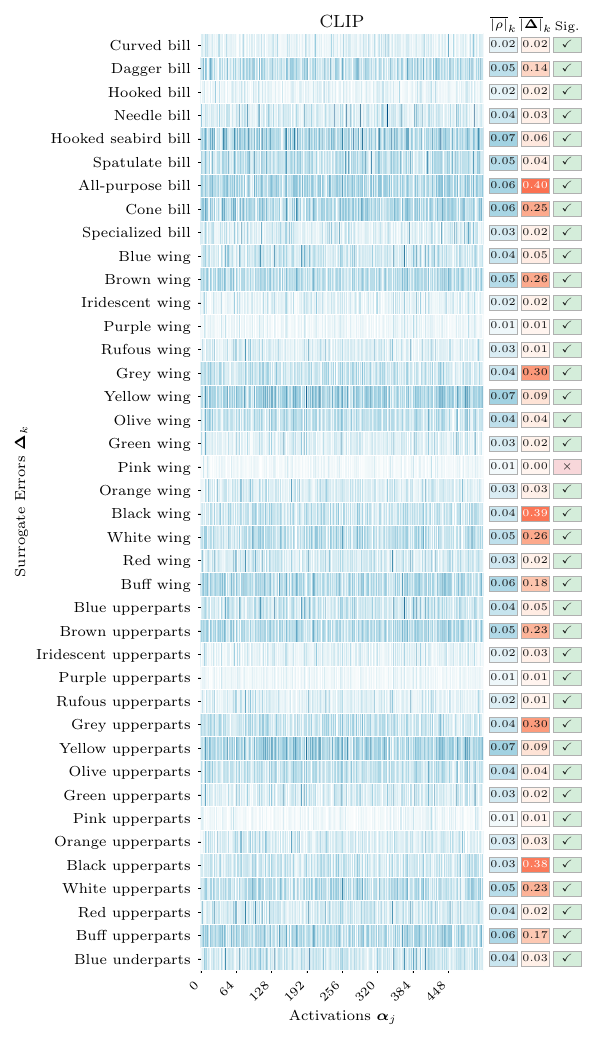}
         \caption{CLIP}
         \label{fig:sys_noise_clip_cub}
     \end{subfigure}
     \hfill
     \begin{subfigure}[t]{0.48\textwidth}
         \centering
         \includegraphics[width=\textwidth,height=0.36\textheight,keepaspectratio]{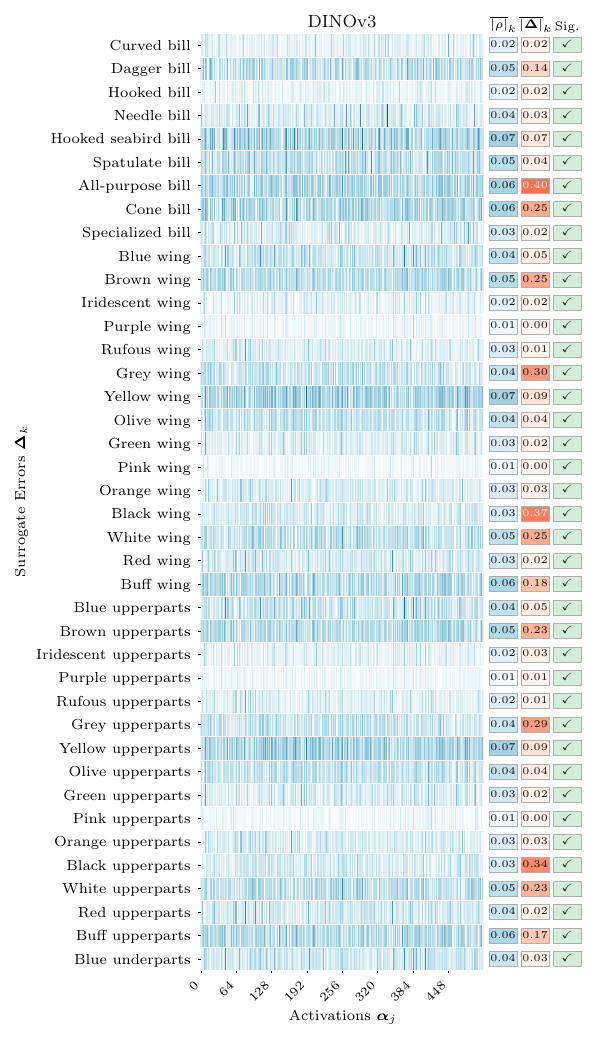}
         \caption{DINOv3}
         \label{fig:sys_noise_dino_cub}
     \end{subfigure}

     \begin{subfigure}[t]{0.48\textwidth}
         \centering
         \includegraphics[width=\textwidth,height=0.36\textheight,keepaspectratio]{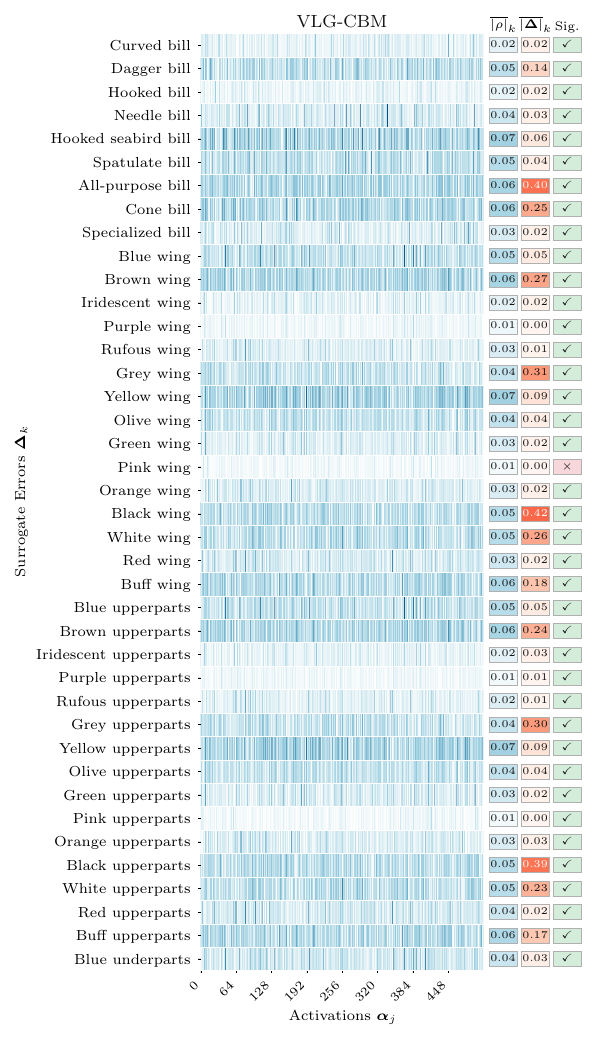}
         \caption{Grounding DINO}
         \label{fig:sys_noise_vlg_cub}
     \end{subfigure}
     \hfill
     \begin{subfigure}[t]{0.48\textwidth}
         \centering
         \includegraphics[width=\textwidth,height=0.36\textheight,keepaspectratio]{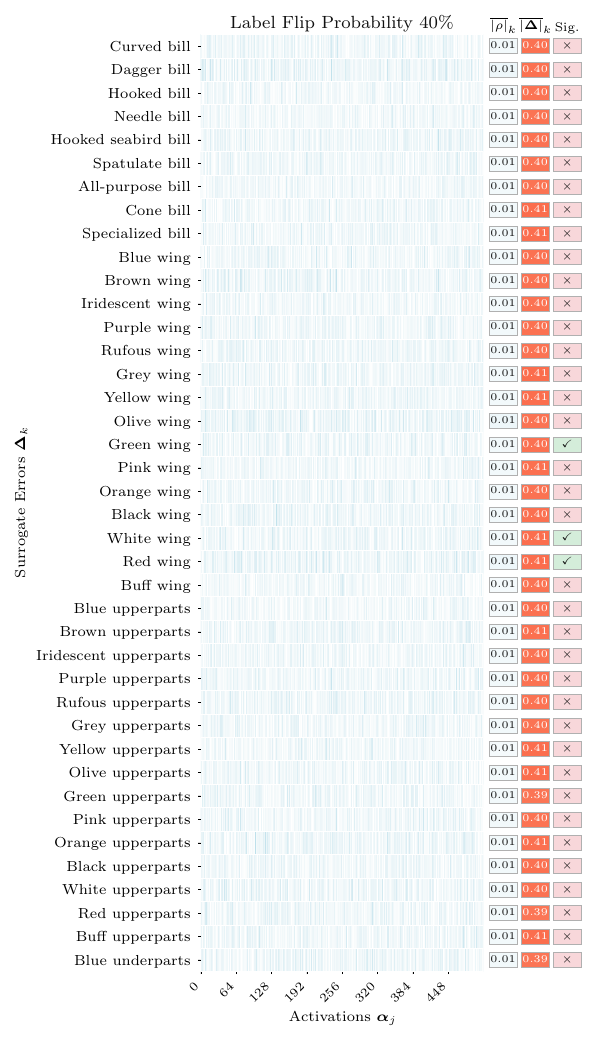}
         \caption{Random Label Noise}
         \label{fig:random_noise_cub}
     \end{subfigure}
     \end{minipage}
     \hfill
     \begin{minipage}{0.10\textwidth}
         \centering
         \includegraphics[width=\textwidth, trim={215 0 0 0}, clip]{figures/pearson_correlation_coefficient_legend.pdf}
     \end{minipage}
    
     \caption{
     Systematic vs. unsystematic surrogate errors for various VLMs (\Cref{apx:vlm-surrogate-functions}) evaluated on the CUB dataset~\cite{welinder2010cub}.
     Following the methodology of \Cref{fig:four_graphs_elements_correlation}, we visualize the Pearson correlation~\cite{pearson1896correlation} between backbone activations and surrogate concept label errors.
     Due to space constraints, we only report results for the first $50$ concepts of the CUB dataset.
     Consistent with our findings for the Elements dataset~\cite{nicolson2025explaining}, the VLM-based surrogate concept labels (a–c) exhibit highly correlated errors, violating the orthogonality condition for faithfulness (\Cref{sec:method-concept-projection}).
     In contrast, the 40\% random noise baseline (d) produces uncorrelated, orthogonal errors. If a concept is predicted without error ($\overline{|\mathbf{\Delta}|}_k = 0$), the correlation coefficient is undefined ($\dagger$).
     }
     \label{fig:four_graphs_cub_correlation}
\end{figure}

\begin{figure}[htbp]
     \centering
     \begin{minipage}{0.86\textwidth}
     \begin{subfigure}[b]{0.45\textwidth}
         \centering
         \includegraphics[width=\textwidth,height=0.36\textheight,keepaspectratio, trim={0 0 0 16}, clip]{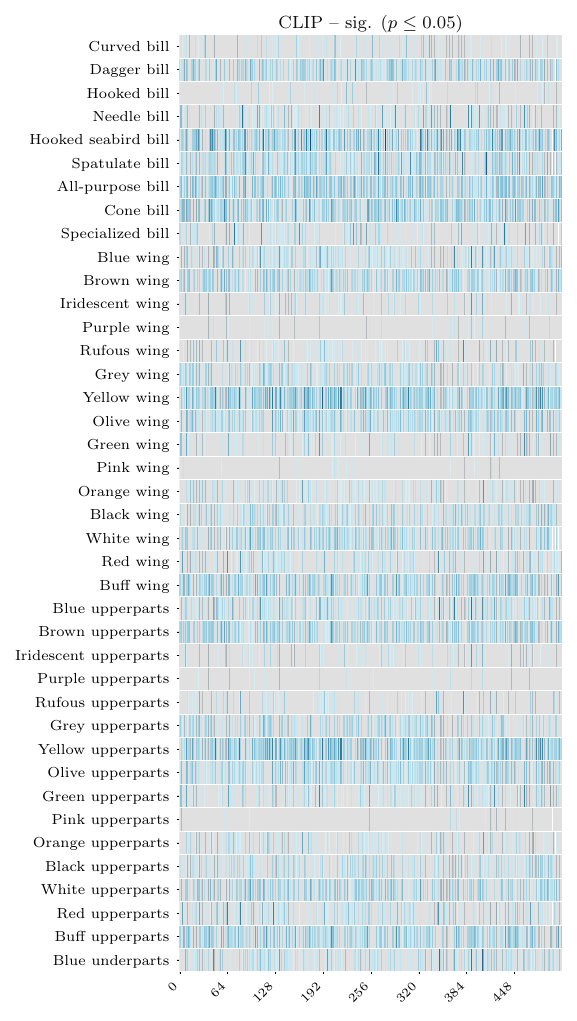}
         \caption{CLIP}
         \label{fig:sys_noise_clip_cub_masked}
     \end{subfigure}
     \hfill
     \begin{subfigure}[b]{0.45\textwidth}
         \centering
         \includegraphics[width=\textwidth,height=0.36\textheight,keepaspectratio, trim={0 0 0 16}, clip]{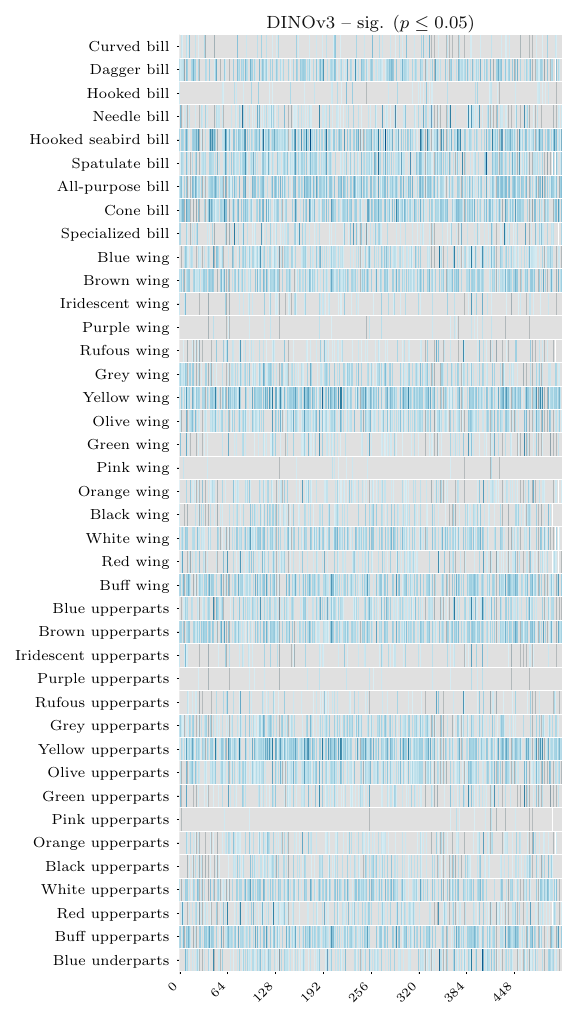}
         \caption{DINOv3}
         \label{fig:sys_noise_dino_cub_masked}
     \end{subfigure}

     \begin{subfigure}[b]{0.45\textwidth}
         \centering
         \includegraphics[width=\textwidth,height=0.36\textheight,keepaspectratio, trim={0 0 0 16}, clip]{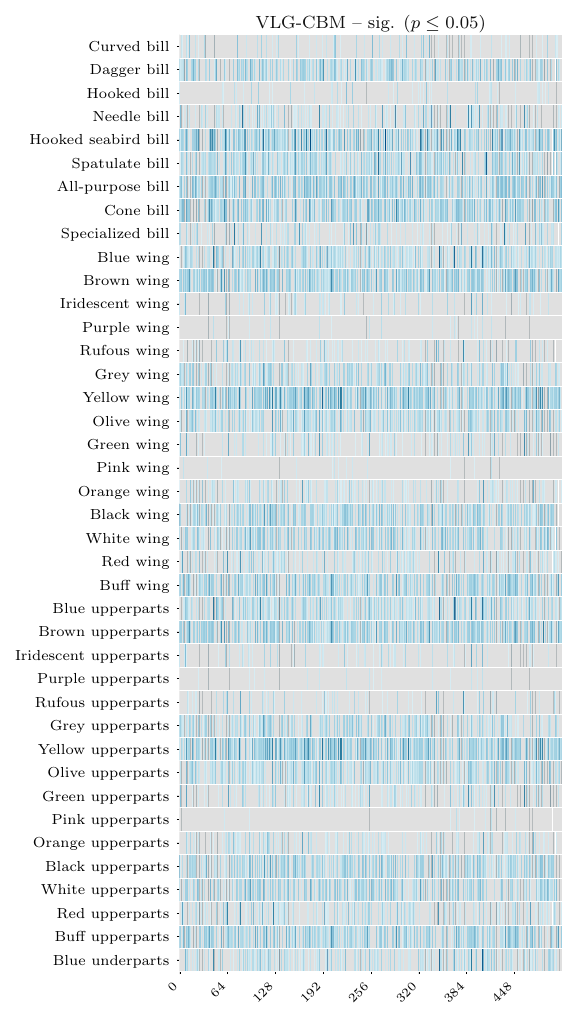}
         \caption{Grounding DINO}
         \label{fig:sys_noise_vlg_cub_masked}
     \end{subfigure}
     \hfill
     \begin{subfigure}[b]{0.45\textwidth}
         \centering
         \includegraphics[width=\textwidth,height=0.36\textheight,keepaspectratio, trim={0 0 0 16}, clip]{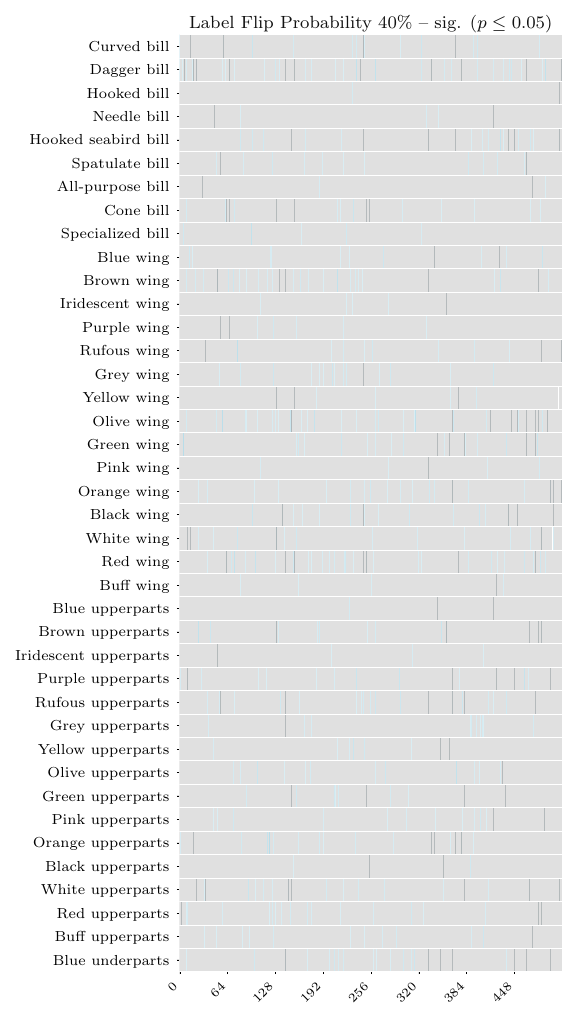}
         \caption{Random Label Noise}
         \label{fig:random_noise_cub_masked}
     \end{subfigure}
     \end{minipage}
     \hfill
     \begin{minipage}{0.10\textwidth}
         \centering
         \includegraphics[width=\textwidth, trim={215 0 0 0}, clip]{figures/pearson_correlation_coefficient_legend.pdf}
     \end{minipage}
     \caption{
     Statistically significant Pearson correlation coefficients~\cite{pearson1896correlation} from \Cref{fig:four_graphs_cub_correlation} (surrogate errors on the CUB dataset~\cite{welinder2010cub}).
     We display only correlation coefficients that remain significant after a Holm-Bonferroni correction~\cite{holm1979correction} ($p < 0.05$).
     The results mirror those of the Elements dataset~\cite{nicolson2025explaining}: the surrogate labels generated by VLM-based methods (a, b, c) produce a dense matrix of statistically significant, correlated errors.
     For the random baseline, we observe a sparse matrix of significant correlations, which matches the expected false-positive rate of repeated testing.
     }
     \label{fig:four_graphs_cub_pvalue}
\end{figure}

\end{appendixblock}
\end{document}